\documentclass[11pt]{article}

\usepackage[preprint]{acl}

\usepackage{times}
\usepackage{latexsym}

\usepackage[T1]{fontenc}

\usepackage[utf8]{inputenc}

\usepackage{microtype}

\usepackage{inconsolata}

\usepackage{graphicx}

\usepackage{hyperref}
\usepackage{url}
\usepackage{booktabs}
\usepackage{tabularx}
\usepackage{graphicx}
\usepackage{multirow}
\usepackage{array}
\usepackage{wrapfig}
\usepackage[many]{tcolorbox}
\usepackage{float}
\usepackage{makecell}
\usepackage{longtable}
\usepackage{amsfonts}

\title{OpenExempt: A Diagnostic Benchmark for Legal Reasoning and a Framework for Creating Custom Benchmarks on Demand}

\author{
    Sergio Servantez$^{1 \dagger}$, 
    Sarah B. Lawsky$^{3}$, 
    Rajiv Jain$^{2}$,
    \\
    \textbf{
    Daniel W. Linna Jr.$^{1}$, 
    Kristian Hammond$^{1}$
    }
    \\
    $^1$Northwestern University, 
    $^2$Adobe Research,
    $^3$University of Illinois Urbana-Champaign\\
    $\dagger$Corresponding author: \texttt{servantez@u.northwestern.edu}
}

\begin{document}
\maketitle

\begin{abstract}
Reasoning benchmarks have played a crucial role in the progress of language models. Yet rigorous evaluation remains a significant challenge as static question-answer pairs provide only a snapshot of performance, compressing complex behavior into a single accuracy metric. This limitation is especially true in complex, rule-bound domains such as law, where existing benchmarks are costly to build and ill suited for isolating specific failure modes. To address this, we introduce OpenExempt, a framework and benchmark for diagnostic evaluation of legal reasoning. The OpenExempt Framework uses expert-crafted symbolic representations of U.S. Bankruptcy Code statutes to dynamically generate a large space of natural language reasoning tasks and their machine-computable solutions on demand. This gives users fine-grained control over task complexity and scope, allowing individual reasoning skills to be probed in isolation. Using this system, we construct the OpenExempt Benchmark, a diagnostic benchmark for legal reasoning with 9,765 samples across nine evaluation suites designed to carefully probe model capabilities. Experiments on 13 diverse language models reveal sharp performance cliffs that emerge only under longer reasoning paths and in the presence of obfuscating statements. We release the framework and benchmark publicly to support research aimed at understanding and improving the next generation of reasoning systems.
\end{abstract}

\begin{figure}[t]
    \centering
    \includegraphics[width=0.9\linewidth]{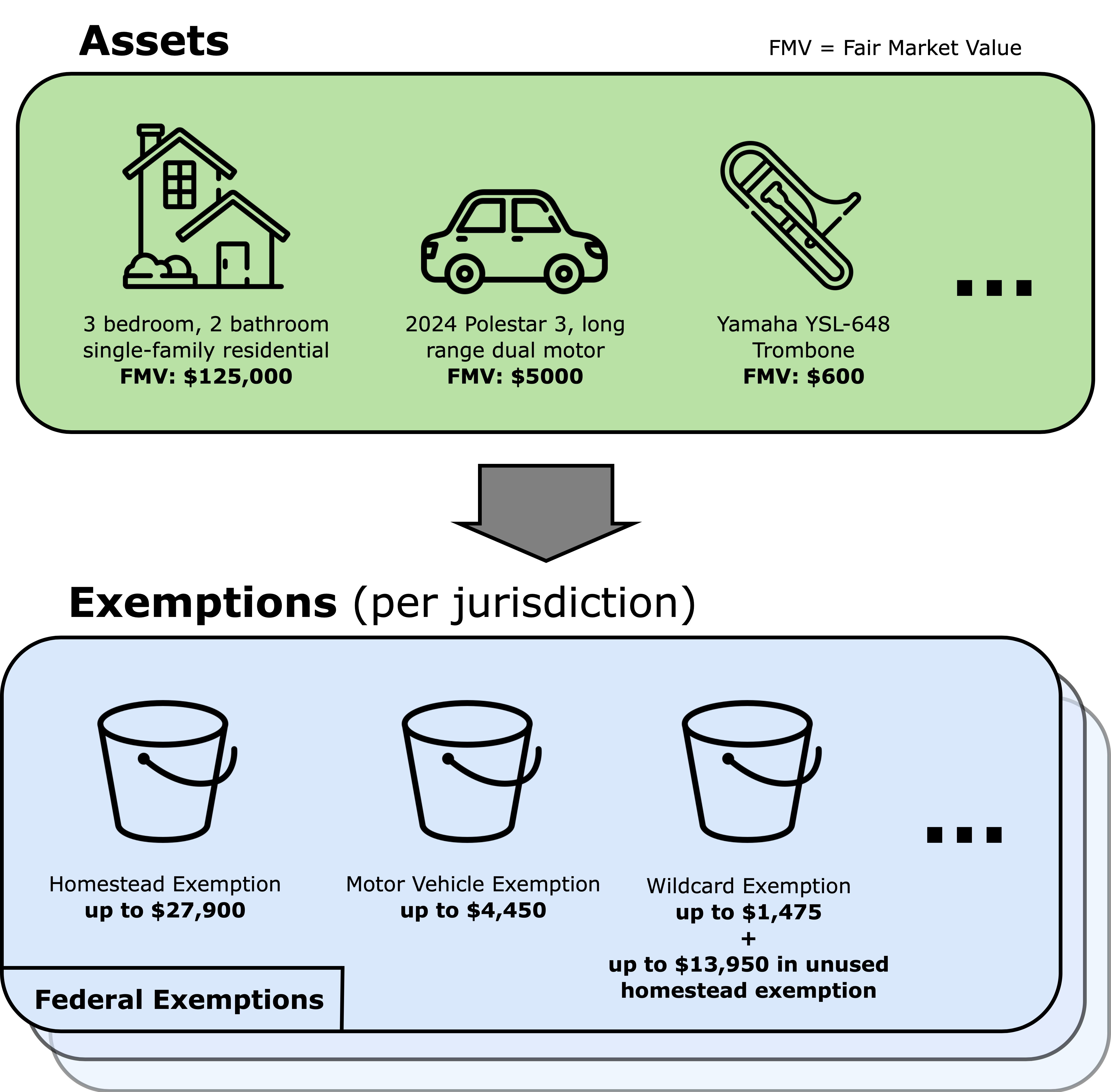}
    \caption[Asset Exemption.]{
        OpenExempt tasks center on U.S. bankruptcy law, primarily asset exemption where assets must be assigned to exemption statutes with dollar limits.
    }
    \label{figure:exemption}
\end{figure}

\section{Introduction}

Language models (LMs) now demonstrate remarkable performance on a wide array of complex tasks, from writing code to passing professional exams. Yet, recent work has begun to question these abilities, probing whether models are truly reasoning or relying on sophisticated forms of memorization and pattern matching \cite{shojaee2025illusionthinkingunderstandingstrengths, mirzadeh2025gsmsymbolicunderstandinglimitationsmathematical}. This uncertainty has fueled a critical need for new evaluation methodologies that move beyond static evaluation \cite{hofmann2025fluidlanguagemodelbenchmarking}, providing deeper, more diagnostic insights into the competencies and failure modes of these powerful systems.

This challenge is particularly acute in the legal domain where reasoning demands precision, consistency, and an understanding of intricate, interdependent rules \cite{servantez2024chainlogicrulebasedreasoning}. Consequently, creating benchmarks for legal tasks has historically relied on expert annotation of solutions, a process that is not only expensive and time-consuming but also results in static datasets of fixed question-answer pairs \cite{10.1093/oxfordhb/9780198940272.013.0007}. Such datasets struggle to keep pace with rapidly evolving models and make it difficult to disentangle the many reasoning skills that a single legal problem may require. A model's failure on a complex task provides only a single, opaque signal of error. Gaining deeper insight requires a more controlled evaluation approach, one where task complexity and scope can be systematically adjusted to reveal a model's specific breaking points.

To address these limitations, we introduce OpenExempt, a framework and benchmark constructed through an interdisciplinary collaboration of computer scientists and legal professionals. The \textbf{OpenExempt Framework} enables fine-grain control over crafting complex legal reasoning tasks and their solutions on demand. This dynamic approach directly overcomes the constraints of static benchmarks by allowing users to vary many aspects of the task, including case details, jurisdictions, and the scope of the task itself, thereby enabling the isolation of different types of reasoning. This makes it possible to disentangle performance across distinct reasoning processes, avoiding the conflation of errors that can obscure a model’s capabilities and limitations.

OpenExempt tasks center on the application of federal and state laws governing the exemption of assets under the United States Bankruptcy Code (U.S. Code Title 11)\footnote{\url{https://uscode.house.gov}}. Inspired by computable contracts where natural language clauses are paired with formal, machine-readable logic \cite{surden2012computable, clack2017smartcontracttemplatesfoundations}, we combine statute text with structured, symbolic representations of their logic and dependencies, making solutions machine-computable. While our approach also requires legal interpretation, OpenExempt does not rely on direct annotation of task solutions. Instead, we use legal knowledge to encode statutes and case assets into structured representations, from which we can construct an immense space of natural language tasks and solutions, removing a key bottleneck in legal reasoning benchmark construction while preserving the accuracy and depth of expert reasoning.

Using this system, we construct the \textbf{OpenExempt Benchmark}, a diagnostic benchmark for legal reasoning composed of nine evaluation suites: 6 diagnostic suites and 3 competency suites. Diagnostic suites isolate specific reasoning challenges by varying task complexity across a single axis, such as the number of assets. This allows us to go beyond single points of failure by precisely measuring the performance delta caused by each variation. Competency suites provide a more holistic assessment of a model's legal reasoning capabilities at three levels of difficulty: basic, intermediate, and advanced.

We present OpenExempt. Our primary contributions are:

\begin{enumerate}
    \item \emph{OpenExempt Framework}: We present a system capable of creating complex legal reasoning tasks and solutions on demand from expert-crafted encodings of statutes and facts. This framework gives users control over defining task scope and complexity, enabling targeted exploration of a vast problem space and diagnostic evaluation through controlled task variation.

    \item \emph{OpenExempt Benchmark}: We introduce a diagnostic benchmark for legal reasoning, consisting of 9 evaluation suites and nearly 10k samples. We perform detailed experiments across a diverse set of 13 language models to surface clear strengths and breaking points.

    \item \emph{Open and Extensible Public Release}: We release the OpenExempt Framework\footnote{\raggedright Code: \url{https://github.com/servantez/OpenExempt}} and Benchmark\footnote{\raggedright Data: \url{https://huggingface.co/datasets/SergioServantez/OpenExempt}} to the public under a permissive license (CC BY 4.0). OpenExempt is intended to support further research in both the legal and NLP communities. Its modular architecture makes it possible for either field to build on this work.
\end{enumerate}

\section{Related Work}

\begin{figure*}[t]
    \centering
    \includegraphics[width=0.9\textwidth]{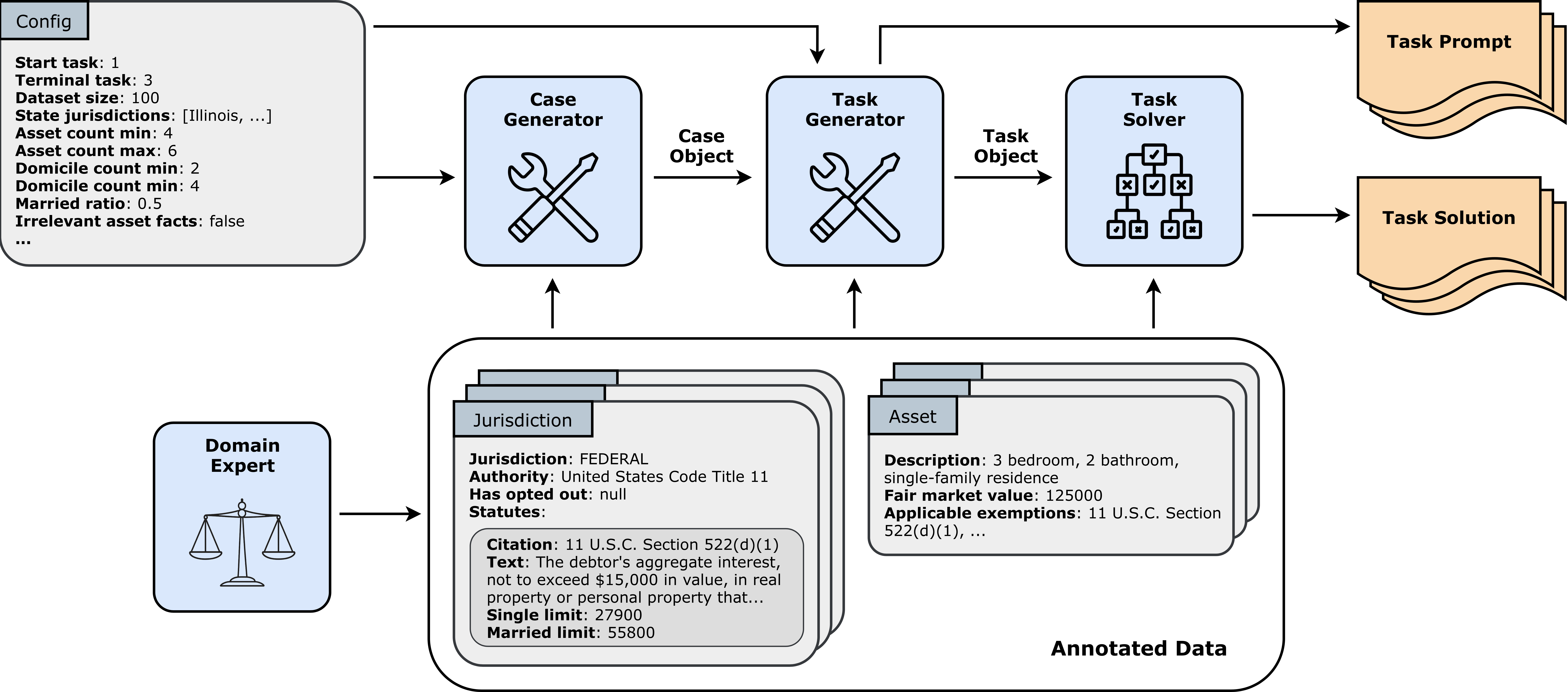}
    \caption[OpenExempt Framework Architecture.]{
        OpenExempt Framework Architecture. Dynamic task generation driven by user-defined configuration and grounded in structured legal knowledge.
    }
    \label{figure:framework}
\end{figure*}

\subsection{Legal Reasoning Benchmarks}
A large body of prior work has examined the legal reasoning capabilities of language models using static datasets of fixed question-answer pairs. Large scale benchmarks like LegalBench \cite{guha2023legalbenchcollaborativelybuiltbenchmark}, LEXTREME \cite{Niklaus_2023}, LawBench \cite{fei2023lawbenchbenchmarkinglegalknowledge}, and LexGLUE \cite{chalkidis2022lexgluebenchmarkdatasetlegal} provide broad assessments across diverse sets of legal tasks. Beyond multi-task benchmarks, other works have targeted specific legal skills, including contract review \cite{hendrycks2021cuadexpertannotatednlpdataset}, legal information retrieval \cite{zheng2025retrieval, joshi2023ucreatunsupervisedcaseretrieval}, legal exam question answering \cite{fan2025lexambenchmarkinglegalreasoning}, case holding identification \cite{zheng2021doespretraininghelpassessing}, legal judgment prediction \cite{chalkidis-etal-2019-neural}, as well as legal datasets for domain adaptation through pretraining \cite{niklaus2024multilegalpile689gbmultilinguallegal, henderson2022pilelawlearningresponsible} and instruction tuning \cite{niklaus2025lawinstructresourcestudyinglanguage}. These benchmarks and datasets have significantly advanced legal reasoning evaluation, yet their static design narrows evaluation to a one-size-fits-all assessment that neither accounts for varying model capabilities nor isolates specific failure modes. OpenExempt introduces a new benchmark paradigm where the user is in control of dynamically crafting legal tasks and defining complexity across multiple dimensions based on their specific evaluation goals.

\subsection{Computable Statutory Reasoning}

Our work is grounded in the field of computational law, where statutes are modeled as executable logic programs. A prominent example is Catala \cite{Huttner2020CatalaMT, lawskyarticle2022}, a domain-specific programming language designed to encode real-world tax laws in an executable form using prioritized default logic \cite{lawskyarticle2017}. Related work has also demonstrated how natural language contracts can be converted into executable programs in the Accord programming language \cite{roche2021ergo}, using an intermediate layer of symbolic legal representations \cite{10.1145/3594536.3595162}. While these works establish the feasibility of symbolic legal encodings, they primarily function as implementation languages or reasoning architectures rather than evaluation benchmarks. SARA \cite{holzenberger2020datasetstatutoryreasoningtax, blairstanek2023gpt3performstatutoryreasoning} is the seminal dataset for evaluating statutory reasoning in language models using Prolog encodings to compute gold solutions for tax problems, yet its approach requires that each scenario be hand-crafted, yielding a fixed dataset of only a few hundred examples. Other prior work has taken an important step toward evaluating hierarchical legal reasoning using case-based analogies, but is also limited by a small, static dataset and does not address statutory reasoning \cite{zhang2025thinkinglongersmarterevaluating}. These threads of work point to the need for a benchmark that is simultaneously dynamic, configurable, diagnostic, grounded in legal knowledge, and scalable beyond hand-crafted datasets – a combination realized in OpenExempt.

\section{OpenExempt Framework}

\begin{figure*}[t]
  \centering
  \includegraphics[width=0.9\textwidth]{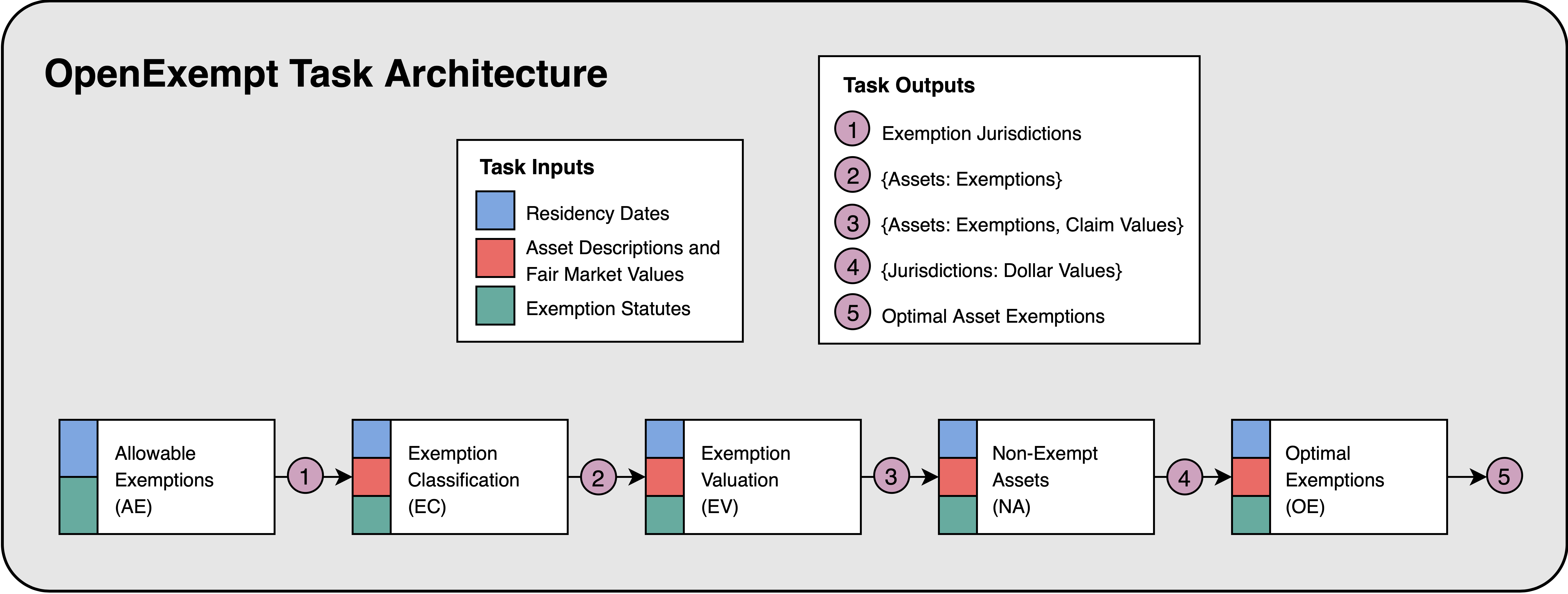}
  \caption[OpenExempt Task Pipeline.]{
  OpenExempt Task Pipeline. Each task depends on the successful completion of its predecessors, forming a composable sequence in which users can select any slice to isolate evaluation of specific reasoning types.
  }
  \label{figure:tasks}
\end{figure*}

The OpenExempt framework is a dynamic task generation system that creates complex legal reasoning benchmarks in the domain of U.S. bankruptcy exemptions. This framework consists of three primary components: 1) a knowledge representation layer that encodes expert legal annotations; 2) a task generator that constructs paired representations in both symbolic and natural language forms; and 3) a deterministic solver that computes ground truth solutions using branch and bound optimization.

The bankruptcy exemption process is an ideal proving ground for controlled evaluation because it allows incremental adjustments to task complexity. Adding assets to a case increases complexity super-linearly: each new asset must not only be evaluated individually, but also in competition with others for shared statutory limits.

\subsection{Asset Exemption in Bankruptcy}

A person filing for bankruptcy, called the Debtor, is allowed to protect certain property from seizure by creditors. An exemption defines a category of property which can be protected - for example, up to \$4,450 in a motor vehicle (Figure \ref{figure:exemption}). Each state enacts its own exemption statutes which differ considerably in regards to which assets are protected. The debtor may claim state or federal exemptions, unless their state specifically prohibits the use of federal exemptions, known as "opt-out"\footnote{11 U.S.C. \S 522(b)(2)}. Which state exemption laws apply to a given case is determined by where the debtor lived in the 730 days before filing for bankruptcy\footnote{Id. \S 522(b)(3)(A)}. Asset exemption is a combinatorial optimization problem, like a legal version of the well known knapsack problem \cite{cormen2022introduction}, where an asset can only be protected by certain exemptions and the goal is to minimize the dollar value of unprotected assets. This process involves many intermediate tasks and can be challenging even for legal professionals.

\subsection{Structured Legal Knowledge}

While the dynamic nature of the framework enables controlled variation in task structure and complexity, it also requires that the underlying legal process be represented in a precise and machine computable form. At the core of OpenExempt are two expert annotated datasets:

\begin{itemize}
    \item \textbf{Debtor Assets}. The framework contains over 500 assets (motor vehicles, real estate, household goods) each manually labeled with the complete set of applicable exemptions for every supported jurisdiction. During task generation, the framework samples from this asset collection, providing one of the many factors that enable OpenExempt to construct a vast combinatorial space of possible cases.
    \item \textbf{Exemption Statutes}. We encode federal and state exemption statutes into structured representations that capture the logical rules required for symbolic reasoning, including monetary caps and state opt-out provisions. We also capture a rich set of constraints and relationships that commonly arise in exemption statutes, as discussed in Section \ref{subsection:constraints}. OpenExempt currently supports federal exemption statutes and state exemptions for Arizona, Illinois, Oregon, Pennsylvania, and Wisconsin\footnote{These states were selected for diversity in both opt-out status and generosity in exemption coverage and limits.}. See Section \ref{appendix:statute_sources} for a list of statute sources.
\end{itemize}

Both datasets adopt a dual representation design, where each asset or statute exists as a pair: a natural language form used in constructing the task prompt, and a structured representation used in computing gold solutions grounded in legal knowledge. This approach is inspired by prior work on smart contracts for legal documents, most notably Accord \cite{roche2021ergo} and Catala \cite{10.1145/3473582}.

\subsubsection{Exemption Constraints and Dependencies} 
\label{subsection:constraints}

\begin{table*}[t]
    \centering
    \small
    \caption{Exemption constraints and dependencies represented in OpenExempt, with citations to exemption statutes that exhibit these properties.}
    \label{table:constraints}
    \begin{tabular}{p{4cm} p{7.3cm} p{3.3cm}}
        \toprule
        \textbf{Variable Name} & \textbf{Description} & \textbf{Example Exemption} \\
        \hline
        \verb|single_limit|, \verb|married_limit|
        & Maximum aggregated dollar amount that may be claimed by a single debtor or a married couple filing jointly.
        & 11 U.S.C. § 522(d)(1)\\
        \hline
        \verb|per_item_limit|
        & Maximum claimable amount per item, distinct from the overall aggregate limit.
        & 11 U.S.C. § 522(d)(3)\\
        \hline
        \verb|single_item_claim_count|, \verb|married_item_claim_count|
        & Restricts the use of an exemption to a single item per claim (e.g., one motor vehicle). Married couples filing jointly may each be entitled to a separate single-item claim (e.g., one motor vehicle each).
        & 735 ILCS 5/12-1001(c)\\
        \hline
        \verb|fallback_exemption|
        & Specifies a relationship with another exemption, whose unused aggregate limit may be reallocated to this exemption.
        & 11 U.S.C. § 522(d)(5)\\
        \hline
        \verb|fallback_single_limit|, \verb|fallback_married_limit|
        & Maximum amount claimable under the fallback exemption, based on marital status.
        & 11 U.S.C. § 522(d)(5)\\
        \hline
        \verb|mutual_exclusion|
        & Defines a mutual exclusion relationship with another exemption, such that claiming either one prohibits the use of the other.
        & Wis. Stat. § 815.18(3)(b)\\
        \bottomrule
    \end{tabular}
\end{table*}

To capture the structure and logic of legal exemptions, such as those found in the U.S. Bankruptcy Code, we introduce a formal representation of exemption constraints and dependencies. These refer to the various common conditions and relationships that govern how an exemption may be applied in practice. Exemption constraints include quantitative or structural limitations, such as caps on the allowable amount per item, differentiated limits for single versus married filers, or restrictions limiting a claim to a single asset. Exemption dependencies, in contrast, encode logical relationships between exemptions, such as mutual exclusions or fallback provisions. Together, these elements form a layer of semantic structure that is critical to accurately modeling exemption behavior and enabling reasoning over exemption applicability and interaction. See Table \ref{table:constraints} for details on exemption constraints and dependencies.

\subsection{Dynamic Task Generation}

OpenExempt dynamically generates benchmark tasks based on a user-defined configuration file that specifies the structure, scope and complexity of the legal problems being created (see Table \ref{appendix:config} for complete list of parameters). This process is largely driven by two components: \texttt{CaseGenerator} and \texttt{TaskGenerator} (Figure \ref{figure:framework}). At runtime, \texttt{CaseGenerator} constructs symbolic bankruptcy cases by sampling case attributes within the bounds set by the configuration. The resulting case object captures relevant legal facts, including parties, marital status, petition date, domicile timeline, and the applicable exemption jurisdiction based on that timeline. \texttt{TaskGenerator} then renders these structured facts into a natural language prompt. When the user-defined task scope includes intermediate subtask solutions, \texttt{TaskGenerator} invokes the OpenExempt task solver to compute the required intermediate outputs and embeds them into the prompt.

To transform structured case data into natural language narratives, OpenExempt uses a template based approach rather than relying on direct end-to-end generation by a language model. While converting structured data into prose is common in other domains, we argue that the precision required for legal text makes unvalidated generation unsuitable, as models frequently introduce subtle ambiguities or inadvertently alter material facts \cite{Dahl_2024}. To address this, we use language models to generate a diverse set of candidate phrasings for narrative elements (e.g., asset ownership), which are then manually screened by a legal professional and converted into parameterized templates. This hybrid approach preserves linguistic variety while ensuring that the resulting fact patterns remain precise and aligned with the ground truth.

\subsection{Computing Gold Solutions}

While OpenExempt tasks are dynamically generated, all solutions are grounded in expert knowledge. The annotated assets enumerate the exemption claims permissible for each asset, while the machine-readable statutes encode the constraints that govern how those claims may be applied. Together, these resources enable the solver to validate candidate outputs and thus define the solution space. For asset-level tasks, like Task EC and EV (defined in Section \ref{subsection:tasks}), the ground truth is directly recovered from the asset annotations for the relevant jurisdiction. For estate-level tasks, like Task NA and OE, which require jointly allocating exemptions across all assets, the framework employs the symbolic solver to perform a branch and bound search over all legally valid exemption assignments. This brute force search is made tractable by pruning partial solutions that cannot surpass the best known allocation, based on remaining exemption capacity and unprocessed assets. Because the solver only explores legally valid allocations, and because all legal rules defining valid claims originate from expert curated encodings, the resulting optimal allocation is both computationally verified and expert grounded.

\subsubsection{Objective Correctness}
\label{appendix:correctness}

Legal reasoning benchmarks must navigate the inherent gray area of statutory interpretation. OpenExempt mitigates this challenge by tightly controlling the scope of legal content and assets included in the benchmark, enabling the construction of tasks where solutions are objectively correct to a high degree of confidence. This requires the deliberate exclusion or modification of statutory provisions that introduce subjectivity. The goal of OpenExempt is not to perfectly model the application of the Bankruptcy Code and state exemption laws, but rather to construct complex legal reasoning tasks with objectively correct answers, which closely resemble real-world legal problems. We prioritize objective correctness through the following design choices:
\begin{itemize}
  \item \textbf{Controlled Asset and Statute Selection}. We curate the pool of assets and exemptions to exclude provisions that rely on subjective standards, such as those requiring an item to be "reasonably necessary". By focusing primarily on tangible assets with clear statutory definitions and avoiding exemptions that depend on complex debtor attributes (disability status, profession), we ensure that the applicability of an exemption is a binary and deterministic question. The description of each asset contains all necessary predicates for a model to determine its eligibility.
  \item \textbf{Normalized Statutory Text for Self-Contained Reasoning}. The exemption statutes in OpenExempt are normalized to eliminate external references and latent ambiguity. For example, an exemption can incorporate requirements defined outside the current title: “Uniforms and accoutrements as provided by 51 Pa.C.S. § 4103”\footnote{42 Pa. Cons. Stat. \S 8124(a)(4)}. In these situations, we omit the reference or inline relevant text if possible. This ensures the model is evaluated on its ability to reason over the task prompt, rather than its ability to recall external legal knowledge not present in that context.
  \item \textbf{Encodable Exemption Logic}. We restrict the benchmark to exemption provisions whose operative logic can be faithfully captured by our formal constraint and dependency representation (Section \ref{subsection:constraints}). While this representation covers many common statutory patterns, not all exemptions can be reduced to these encodings given the infinite variability in natural language. By excluding provisions that cannot be encoded, we ensure that our derived ground truth solutions remain computationally verifiable.
\end{itemize}

\section{OpenExempt Benchmark}

\begin{figure}
    \centering
    \includegraphics[width=\linewidth]{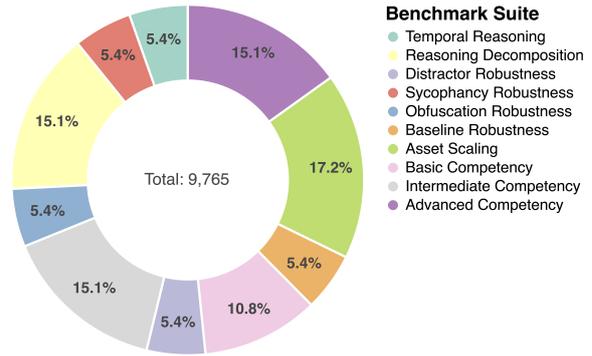}
    \caption[Benchmark Suite Distribution.]{
        Sample distribution across benchmark suites.
    }
    \label{figure:suite_distribution}
\end{figure}

Using the above framework, we construct the OpenExempt benchmark consisting of 9,765 samples across nine evaluation suites (Figure \ref{figure:suite_distribution} shows the sample distribution across suites).

\subsection{Tasks}
\label{subsection:tasks}

\begin{figure*}[t]
  \centering
{
\small
\begin{tcolorbox}[enhanced,
  attach boxed title to top center={yshift=-3mm,yshifttext=-1mm},
  colback=white,
  colframe=blue!50!black,
  colbacktitle=blue!50!black,
  title=Task EC Example: Distractor Robustness Suite,
  fonttitle=\bfseries
  ]
    For each asset in the estate, identify all applicable exemptions under which that asset may be protected.
    \\\\
    Your answer to this task must be based solely on applying the provided Federal and State statutes to the given facts. If the task involves a married couple, assume all assets mentioned are jointly owned, with each spouse holding an equal undivided interest, unless explicitly stated otherwise. Assume all assets are held for the personal use of the Debtor(s), unless explicitly stated otherwise.
    \\\\
    Facts:
    Megha and Dalia Joshi (hereinafter the Debtors) filed for bankruptcy on March 7th 2024. The Debtors began living in Litchfield Park, Arizona on 19 March 2012, but relocated to Waunakee, Wisconsin on 9.10.2021. For the 28 days following that date, Megha stayed in Patagonia, Arizona to complete a boater safety course and obtain a state-issued boating certificate, a new requirement for their job as a marine biologist. While there, they signed a rental agreement (Contract \#R-781) for a boat slip at a local marina to berth the assigned training vessel for the duration of the course. The Joshis chose to relocate their household to Weyauwega, Wisconsin on 12th of September, 2021. Records show that Megha and Dalia Joshi possess a compact Bluetooth speaker with splash-resistant casing with a value of \$400.00. Megha and Dalia own a calico cat worth \$145.00. The Debtor's name is listed on a UTMA savings account, and the corresponding Form 1099-INT is mailed to their address. The account holds a balance of \$8,150, which originated as an irrevocable gift from the Debtor's brother to the account's beneficiary, the Debtor's 14 year-old nephew. The known assets of Megha and Dalia Joshi include a woven tapestry wall hanging with bohemian motif worth \$1,305.00 and a Hi-Point C9 9mm pistol appraised at \$195.00.
\end{tcolorbox}
}
  \caption{Example Task EC (Exemption Classification) prompt from the Distractor Robustness evaluation suite; response format instructions and statutes abridged for brevity.}
  \label{figure:prompt}
\end{figure*}

OpenExempt is composed of five tasks, with a total of 15 task variants, that mirror the sequence of legal reasoning steps a debtor’s attorney performs when protecting assets in bankruptcy. For each task, the model receives a fact pattern detailing the debtor's situation, which may include asset disclosures and residential history, along with a corpus of relevant federal and state laws. We show an example Task EC prompt in Figure \ref{figure:prompt}, provide additional prompt examples with solutions in Section \ref{appendix:task_examples}, and describe each task below:

\begin{itemize}
  \item \textbf{Task AE (Allowable Exemptions):} Before exemptions can be claimed, the Bankruptcy Code requires first understanding which state or federal exemptions are available to the Debtor. This task involves applying the multi-step "730-day Rule"\footnote{11 U.S.C. \S 522(b)(3)(A)} to the Debtor's residency history to identify the applicable exemption jurisdictions, while accounting for state opt-out provisions\footnote{Id. \S 522(b)(2)}.
  \item \textbf{Task EC (Exemption Classification):} Once the allowed exemption jurisdictions have been identified, each asset must be matched to the categories of exempt property defined by statute. This task requires rule-based reasoning to determine if a given asset satisfies the exemption antecedent, the specific property category defined by the statute. This is a multi-label classification problem since multiple exemptions can apply to a single asset.
  \item \textbf{Task EV (Exemption Valuation):} Exemptions are typically limited to a fixed dollar amount defined by the statute. This task requires not only identifying applicable exemptions, but also applying these statutory caps to calculate the maximum protectable dollar value for each asset under each of its available exemptions. Tasks EC and EV require asset-level reasoning since each asset-exemption pair is considered independently, without factoring in aggregate limits.
  \item \textbf{Task NA (Non-exempt Assets):} Tasks NA and OE demand estate-level reasoning to strategically allocate exemptions across all the Debtor's assets. This task requires solving this strategic allocation to determine the minimal total dollar value of non-exempt assets after applying all applicable exemptions, for each allowable exemption jurisdiction.
  \item \textbf{Task OE (Optimal Exemptions):} This task requires articulating the complete, optimal strategy to achieve the best outcome from Task NA. This requires selecting the allowable exemption jurisdiction that minimizes non-exempt asset value, and generating the explicit exemption schedule for that optimal jurisdiction. The schedule must produce a complete mapping of which exemptions, and what dollar amounts, are allocated to each specific asset.
\end{itemize}

\begin{table*}[t]
    \centering
    \caption{Model performance (sample-based $F1$) by task across Basic (bc), Intermediate (ic), and Advanced Competency (ac) suites.}
    \label{tab:competency}
    \small
    \setlength{\tabcolsep}{2.5pt} 
    \begin{tabularx}{\textwidth}{@{}l*{5}{>{\centering\arraybackslash}X}@{}}
        \toprule
        \textbf{Task}  & AE & EC & EV & NA & OE \\
        \cmidrule(lr){2-2}\cmidrule(lr){3-3}\cmidrule(lr){4-4}\cmidrule(lr){5-5}\cmidrule(lr){6-6}
        \textbf{Suite} & bc / ic / ac & bc / ic / ac & bc / ic / ac & bc / ic / ac & bc / ic / ac \\
        \midrule
        GPT-5 & .884 /.743 /.612 & \textbf{.924} /\textbf{.744} /.554 & .893 /.635 /.496 & .893 /\textbf{.733} /\textbf{.558} & .933 /\textbf{.744} /.404 \\
        o3 & .917 /.747 /.604 & .901 /.743 /\textbf{.571} & \textbf{.912} /\textbf{.651} /.500 & \textbf{.949} /.728 /.548 & \textbf{.944} /.724 /.408 \\
        o4-mini & .940 /\textbf{.757} /.621 & .711 /.575 /.418 & .691 /.541 /.388 & .744 /.539 /.326 & .844 /.543 /.265 \\
        Claude-Sonnet-4 & .940 /.743 /.625 & .723 /.502 /.452 & .697 /.478 /.353 & .718 /.499 /.340 & .805 /.476 /.255 \\
        Gemini-2.5-Pro & .943 /.753 /.605 & .900 /.740 /.549 & .877 /.623 /\textbf{.502} & .889 /.665 /.540 & .938 /.714 /\textbf{.518} \\
        DeepSeek-R1 & \textbf{.957} /.728 /.610 & .809 /.612 /.457 & .771 /.546 /.364 & .860 /.649 /.476 & .901 /.607 /.347 \\
        \midrule
        GPT-4.1 & .955 /.714 /.588 & .522 /.276 /.224 & .453 /.207 /.163 & .689 /.504 /.316 & .777 /.538 /.240 \\
        Llama-4-Maverick & .865 /.659 /.586 & .515 /.317 /.226 & .496 /.320 /.193 & .554 /.227 /.122 & .703 /.260 /.095 \\
        DeepSeek-V3 & .942 /.673 /\textbf{.627} & .598 /.365 /.291 & .530 /.366 /.274 & .594 /.393 /.196 & .802 /.330 /.170 \\
        \midrule
        Claude-3.5-Haiku & .707 /.572 /.360 & .529 /.411 /.317 & .415 /.344 /.256 & .502 /.204 /.069 & .561 /.147 /.026 \\
        Gemma-3 & .710 /.539 /.403 & .503 /.447 /.331 & .373 /.264 /.191 & .404 /.224 /.137 & .660 /.140 /.007 \\
        Gemini-2.5-Flash & .935 /.723 /.574 & .835 /.671 /.474 & .836 /.562 /.396 & .870 /.586 /.441 & .935 /.671 /.355 \\
        Llama-4-Scout & .596 /.480 /.386 & .401 /.360 /.281 & .350 /.294 /.172 & .422 /.188 /.118 & .569 /.136 /.033 \\
        \bottomrule
    \end{tabularx}
\end{table*}

\textbf{Task Variants}. Since OpenExempt tasks form a sequential pipeline (Figure \ref{figure:tasks}), solving any given task depends on the successful completion of its predecessors. The OpenExempt framework allows users to configure which earlier steps are already solved and provided in the task prompt. This creates a family of task variants, where a task can be presented in its vanilla form (no prior steps solved) or with some or all preceding steps solved by the framework. Each variant corresponds to a contiguous interval of the pipeline. This design enables a fine-grained assessment of how cumulative complexity and error propagation impact model performance, a capability we later demonstrate with the Reasoning Decomposition evaluation suite.

\subsection{Benchmark Suites}

The OpenExempt Benchmark organizes its evaluation into nine suites designed to capture both broad and fine-grained assessments of legal reasoning. The three competency suites evaluate a wide range of exemption scenarios to provide a holistic view of a model’s reasoning capabilities, mirroring a traditional benchmark. In contrast, the six diagnostic suites isolate and vary one specific dimension of task complexity, such as the density of obfuscating statements. This approach enables targeted, causal analysis, allowing us to isolate and understand precisely which specific reasoning challenges are contributing to performance degradation. See Table \ref{appendix:suite_config} for a summary of configuration settings for each suite.

\subsubsection{Competency Suites}

\textbf{Basic, Intermediate, and Advanced Competency}. These three suites form a structured progression of difficulty where higher tiers contain more complex fact patterns, including larger asset pools, more extensive domicile histories, and exemption statutes drawn from a broader set of state jurisdictions. This tiered design ensures that the benchmark remains informative across a wide range of model capabilities: smaller models can be meaningfully evaluated on the lower tiers without collapsing to failure, while larger models can be challenged at higher tiers without saturating performance. In this way, the competency suites yield reliable, discriminative signals that align with the reasoning capacity of the model being assessed.

\begin{figure*}
    \centering
    \includegraphics[width=\linewidth]{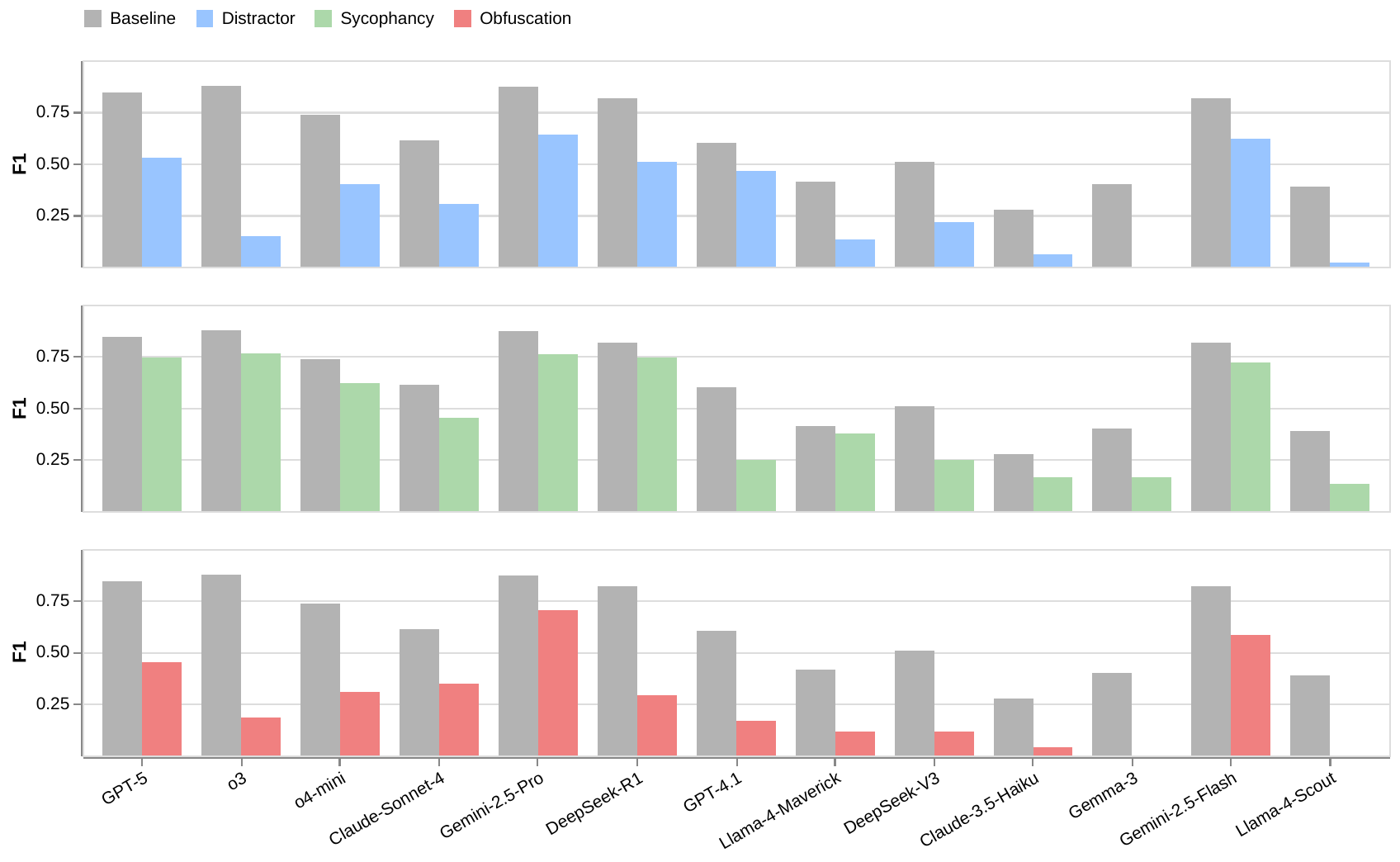}
    \caption[Model Performance (F1) on Task OE under Distractor, Sycophancy, and Obfuscation Perturbations.]{
        Model performance (F1) on Task OE under Distractor, Sycophancy, and Obfuscation perturbations. Colored bars show performance under each robustness suite, highlighting the degree to which model accuracy degrades relative to baseline.
    }
    \label{figure:or}
\end{figure*}

\subsubsection{Diagnostic Suites}

\textbf{Temporal Reasoning}.
This suite isolates reasoning about temporal rules that govern exemption eligibility under the Bankruptcy Code. In these tasks, the Debtor’s prior residences are spread across multiple states and dates, and the model must determine which exemptions the Debtor is permitted to claim by correctly applying the 730-day Rule under 11 U.S.C. \S 522(b)(3)(A). This rule is a three part statutory test that requires reasoning about both the duration and location of the Debtor’s domicile. By increasing the complexity of the domicile history while holding all other settings constant, we can precisely measure how temporal complexity affects model performance.

\textbf{Reasoning Decomposition}. This diagnostic suite measures the effects of cumulative complexity and error propagation across the OpenExempt task pipeline (Figure \ref{figure:tasks}). It evaluates Tasks EC, EV, NA and OE, by testing against all possible preceding task variants (Task AE has no preceding tasks). This configuration allows for a causal decomposition of total error into two components: stage error, which reflects the model's inability to solve the target task itself (e.g., Task EV), and propagation error, which arises from the model's reliance on its own incorrect conclusions from preceding steps (e.g., Tasks AE and EC).

\textbf{Distractor, Sycophancy, and Obfuscation Robustness}.

The OpenExempt benchmark includes three diagnostic suites designed to evaluate how well models maintain legal accuracy when faced with extraneous, misleading, or irrelevant information within the fact pattern. OpenExempt supports two forms of obfuscation: irrelevant facts, which introduce legally immaterial details about assets or prior residences (e.g., property not owned by the Debtor or travel that does not alter domicile), and opinions, which present subjective statements about assets or exemption eligibility that carry no legal force. The Distractor Robustness suite introduces irrelevant facts, testing whether models can ignore seemingly pertinent but legally inconsequential information. The Sycophancy Robustness suite introduces opinion statements, measuring whether models are influenced by subjective assertions rather than statutory requirements. The Obfuscation Robustness suite combines both types, presenting the full range of distracting and misleading content. The effect of each perturbation is isolated by comparison to a baseline configuration with no obfuscating statements. Figure \ref{figure:prompt} shows an example task prompt with embedded distractor statements.

\textbf{Asset Scaling}.

Asset pool size is a significant driver of task complexity. This suite evaluates how model performance changes as the number of assets in the Debtor’s estate is incrementally scaled. The primary source of emerging difficulty is the strategic competition between assets for limited statutory exemption values. In cases with one or two assets, the optimal allocation of exemptions can be trivial. However, as the asset pool size grows and multiple assets become eligible for the same finite exemption values, the task transitions into a complex optimization problem. The model is forced to consider alternative exemptions and strategically allocate the available dollar value of each to maximize the total protected value of the estate, thereby demanding a much stricter and more comprehensive legal reasoning process.

\begin{figure*}[t]
    \centering
    \includegraphics[width=\linewidth]{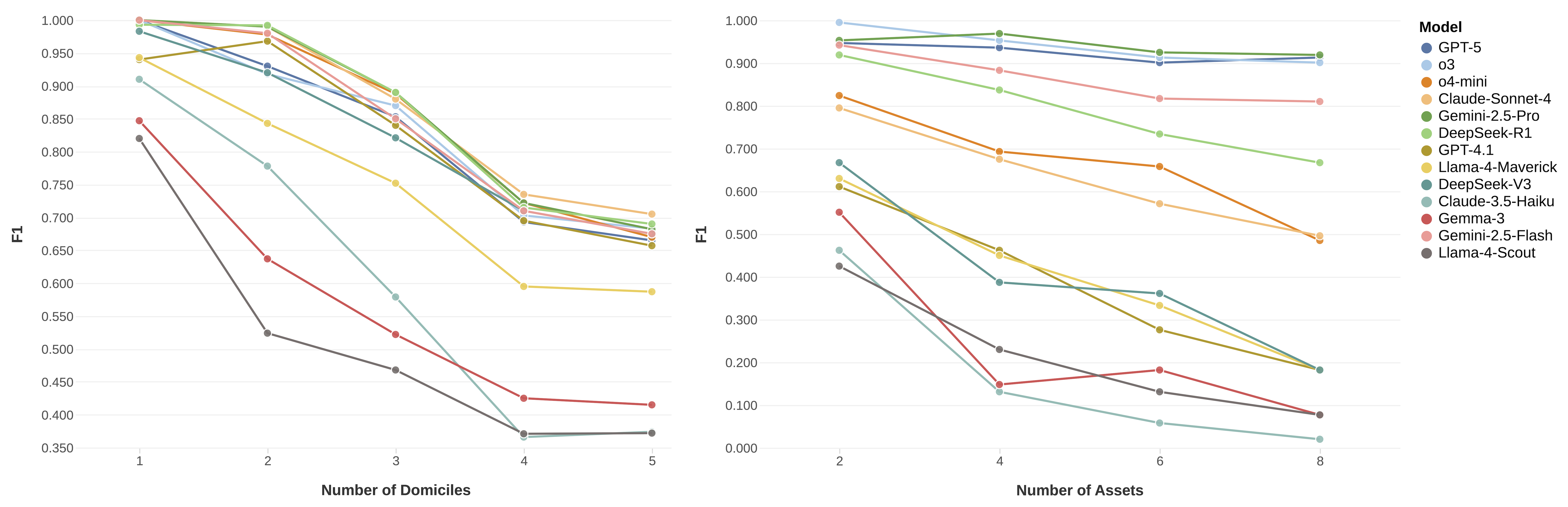}
    \caption[Model performance (F1) on Temporal Reasoning (left) and Asset Scaling (right) suites.]{
        Model performance (F1) on Temporal Reasoning (left) and Asset Scaling (right) suites.
    }
    \label{figure:tr_as}
\end{figure*}

\section{Results}

We summarize our findings here, and provide complete experiment details in the \hyperref[appendix:section]{Appendix}.

\subsection{Experimental Setup}

To support few-shot learning, we follow LEXam \cite{fan2025lexambenchmarkinglegalreasoning} and split 5 samples from each task dataset into a dev set, with the remaining 100 samples in the test set. Evaluation suites contain a collection of these 105-sample datasets, each with its own configuration file. Across all suites, this yields 9,765 samples in total, split into 9,300 test samples and 465 dev samples. Prior work has shown that language models can struggle with in-context demonstrations in the legal domain \cite{servantez2024chainlogicrulebasedreasoning}. Therefore, we focus this work on evaluating models in a zero-shot setting to establish baseline performance, but leave exploration of few-shot learning for future work.

\subsection{Models}
We evaluate 13 language models grouped into three categories: 1) \textit{reasoning models}: GPT-5 \cite{openai2025gpt5}, Claude-Sonnet-4 \cite{anthropic2025sonnet4}, Gemini-2.5-Pro \cite{google2025geminipro}, o3 \cite{openai2025o3o4mini}, o4-mini \cite{openai2025o3o4mini}, and Deepseek-R1 \cite{deepseekai2025deepseekr1incentivizingreasoningcapability}; 2) \textit{large models}: GPT-4.1 \cite{openai2025gpt41}, Llama-4-Maverick (17B-128E-Instruct) \cite{meta2025llama4}, and Deepseek-V3 \cite{deepseekai2025deepseekv3technicalreport}; 3) \textit{efficient models}: Gemini-2.5-Flash \cite{google2025geminiflash}, Claude-3.5-Haiku \cite{anthropic2024claude3}, Llama-4-Scout (17B-16E-Instruct) \cite{meta2025llama4}, and Gemma-3-(27b-it) \cite{gemmateam2025gemma3technicalreport}. We use a temperature of 0 for all models that support temperature, except for DeepSeek-R1 which we set to 0.6 based on developer recommended settings \cite{deepseek2025r1}. We set max token length to 16384, or the maximum token length supported by the model if it is less. We find these extended outputs are necessary to ensure complete answers.

\subsection{Evaluation Protocol}

Across all tasks, OpenExempt reports precision, recall and F1 scores computed at the sample level and then macro-averaged across samples. For asset-level tasks (EC, EV), the evaluator first computes per asset scores within a case, then averages across assets to determine the sample score, preventing assets with more applicable exemptions from dominating the aggregate. For tasks with multi-label predictions (AE, EC, EV), we evaluate using set overlap across discrete labels (jurisdictions or exemption citations). For tasks involving dollar valued predictions (EV, NA, OE), we additionally compute mean absolute relative error (MARE) between predicted and gold amounts. A numeric prediction is treated as correct if it falls within a 5\% absolute relative error tolerance of the corresponding gold value. Because relative error can become unstable for gold amounts near zero, we add a small stabilizing constant $\epsilon$. Formally, for a predicted amount $\hat{y}$ and gold amount $y$, we define the within-tolerance indicator:
\[
\mathbb{I}_{\tau}(\hat{y}, y)
=
\mathbf{1}\!\left[\left|\frac{\hat{y}}{y+\epsilon}-1\right|<\tau\right]_{\epsilon=1,\;\tau=0.05}.
\]

For tasks that require structured outputs (EC, EV, NA, OE), predictions that fail to parse are marked as \textit{invalid format} and are scored as incorrect, making format compliance a measured component of the benchmark. We observe a low rate of malformed responses across all evaluated models, typically below 1\% (Table \ref{appendix:malformed_response}). We discuss LM response validation in detail in Section \ref{appendix:response_validation}.

\subsection{Competency Evaluation}

\begin{figure}[t]
    \centering
    \includegraphics[width=\linewidth]{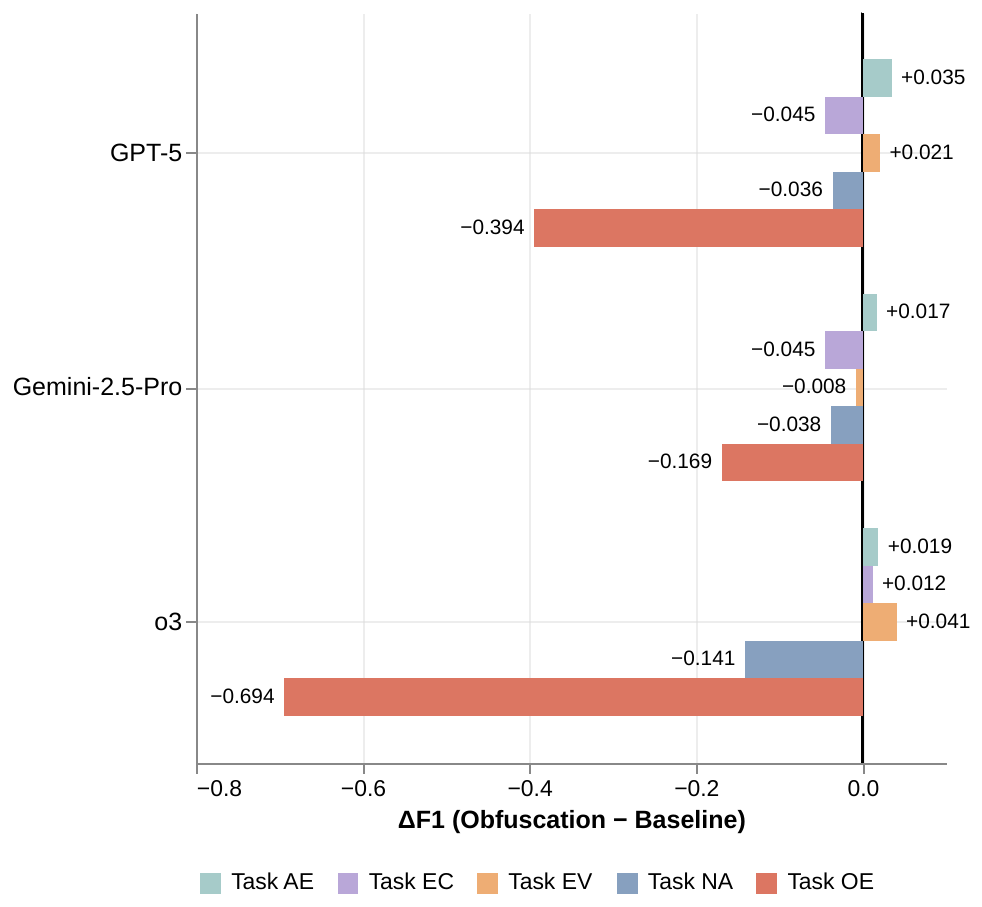}
    \caption[Reasoning Decomposition on Task OE.]{
        Obfuscation Robustness by task for three top performing reasoning models. Each bar shows the absolute change in F1 ($\Delta F_1$) under obfuscation perturbations, computed as obfuscation minus baseline. Obfuscating statements are identical across tasks. Positive values indicate performance gains, negative values indicate degradation.
    }
    \label{figure:or_delta}
\end{figure}

\textit{Basic, Intermediate, and Advanced Competency}. \textbf{Competency suites produce stable, monotonic performance degradation across increasing difficulty levels, yielding clear and reliable distinctions between model reasoning capacities} (Table~\ref{tab:competency}). Reasoning LMs predominantly outperform efficient and large non-reasoning models, with performance gaps particularly pronounced in more difficult settings. Gemini-2.5-Flash is a clear exception to this trend: despite being an efficient model, it consistently performs closer to reasoning models across our evaluations. GPT-5 and Gemini-2.5-Pro often rank among the best performing models. Yet, model recency alone does not explain performance, as o3 frequently outperforms newer models, even in the advanced tier. Notably, no model achieves an F1 score above 0.63 on any task in the advanced suite, indicating room for improvement in advanced legal reasoning where multi-step inference and complex rule application are prevalent. The Basic Competency suite reveals clear distinctions between efficient models on Task OE that vanish under advanced settings. For example, Gemma-3 outperforms Llama-4-Scout in the basic tier ($F1$ = 0.66 vs. 0.569), but both models collapse to near-zero performance in the advanced tier. Conversely, reasoning models approach saturation on simpler tasks ($F1>0.9$ on AE-bc), masking capabilities that only diverge under higher complexity settings. Our tiered evaluation approach mitigates these floor and ceiling effects that would otherwise obscure distinctions between models. To disentangle the failures observed in the competency evaluations, we next turn to the diagnostic suites.

\subsection{Diagnostic Evaluation}

\begin{figure}[t]
    \centering
    \includegraphics[width=\linewidth]{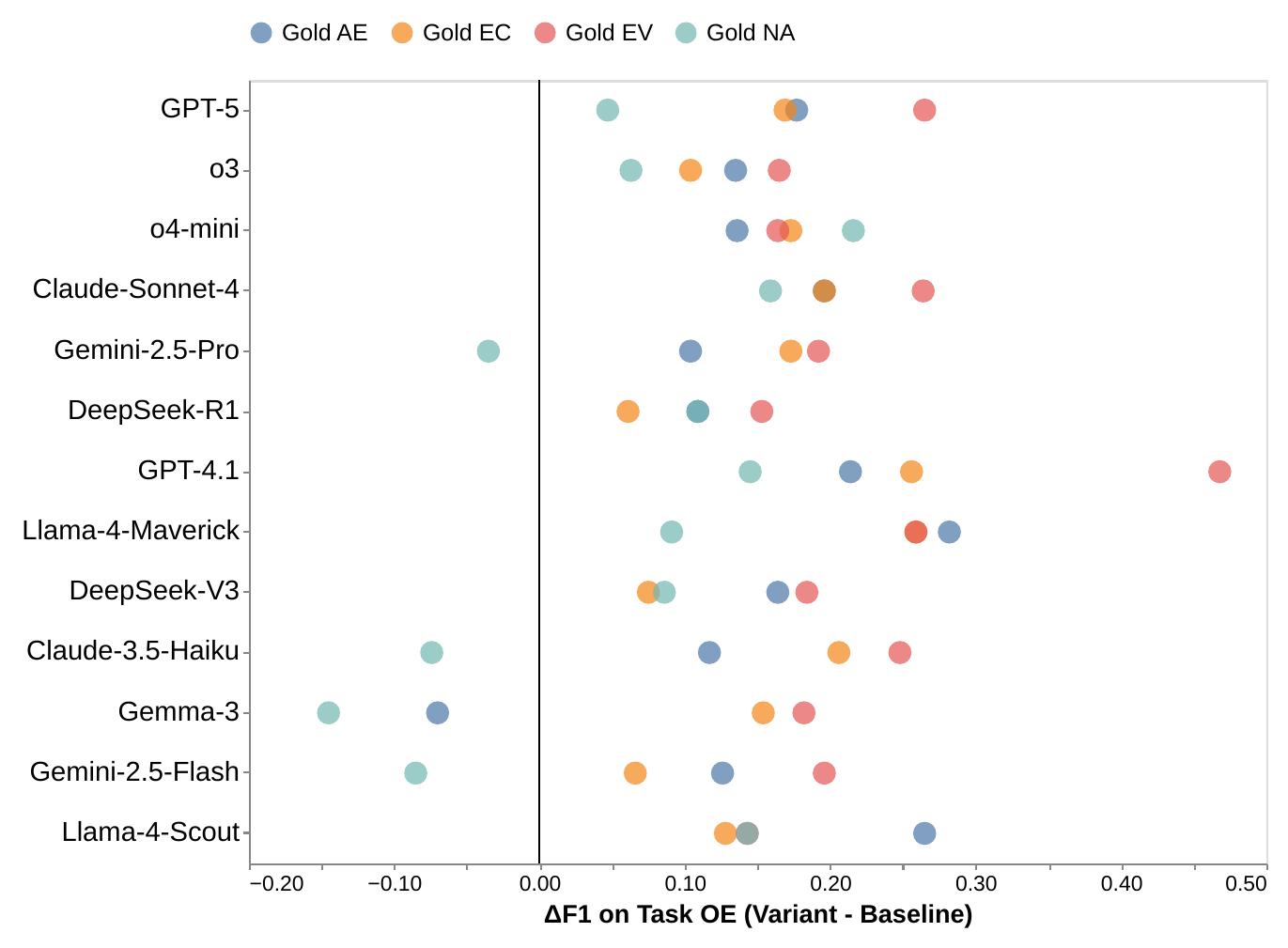}
    \caption[Reasoning Decomposition on Task OE.]{
        Reasoning Decomposition on Task OE. Each point shows the absolute change in F1 ($\Delta F_1$) when a model is provided with gold solutions to a specific intermediate subtask, computed as variant minus baseline (no solved steps). Positive values indicate performance gains, negative values indicate degradation.
    }
    \label{figure:rd}
\end{figure}

\textit{Temporal Reasoning}. \textbf{Temporal reasoning fails at a predictable threshold for reasoning models: performance declines modestly, then drops sharply, before leveling off} (Figure \ref{figure:tr_as}, Table \ref{appendix:tr}). As temporal complexity increases while holding all other settings constant, reasoning models achieve perfect or near perfect performance with one domicile, and degrade only slightly at two (typically less than 0.03 $F1$). Performance drops become noticeably sharper as complexity reaches three and four domiciles, with four marking the clearest breaking point (at least a 0.145 $F1$ decrease for all reasoning models). This trend does not continue at five domiciles, where performance decreases typically return to only a few $F1$ points. We observe a similar pattern across all reasoning models (Figure \ref{figure:tr_as}). In contrast, efficient models exhibit greater variability, and tend to degrade earlier and more smoothly as domicile complexity increases.

\textit{Asset Scaling}. \textbf{Asset scaling sharply separates model capacity: a select few frontier reasoning models remain robust under multi-asset exemption optimization, while many other models collapse.} (Figure \ref{figure:tr_as}, Tables \ref{appendix:as_efficient} to \ref{appendix:as_reasoning}). Performance declines are comparatively modest on asset-level tasks (EC, EV), even for efficient models, but become markedly sharper on estate-level tasks (NA, OE). For example, when scaling assets from 2 to 8, the $F1$ score for Llama-4-Scout decreases by 0.057 (EC) and 0.053 (EV), compared to 0.177 (NA) and 0.348 (OE). This gap reflects the shift from local exemption decisions to globally constrained allocation across competing assets. However, this degradation pattern is not universal, as three reasoning models (GPT-5, o3, Gemini-2.5-Pro) prove significantly more robust to increases in asset count, even where optimization demands are most acute. For example, GPT-5 declines by only 0.034 $F1$ on Task OE under full asset scaling (2 to 8 assets). These results indicate that the performance drops observed in the Advanced Competency suite cannot be attributed to asset complexity alone, but rather to the interaction of multiple difficulty dimensions.

\textit{Distractor, Sycophancy, and Obfuscation Robustness}. \textbf{Model vulnerability to obfuscation is not uniform: identical statements can produce disparate levels of harm, which compound as legal reasoning becomes more complex.} (Figures \ref{figure:or} and \ref{figure:or_delta}, Tables \ref{appendix:or_efficient} to \ref{appendix:or_reasoning}). Model performance typically declines in the presence of obfuscating statements (distractors, opinions), but several models exhibit slight performance increases on simpler tasks (AE, EC), particularly under distractors alone. This pattern suggests extraneous information can encourage more deliberate analysis in less complex settings. Across Tasks AE through NA, the strongest reasoning models (GPT-5, o3, Gemini-2.5-Pro) exhibit little to no degradation under any obfuscation setting (Figure \ref{figure:or_delta}). Yet for Task OE, all reasoning models show substantial declines, with distractor and obfuscation perturbations producing the sharpest drops (Figure \ref{figure:or} and \ref{figure:or_delta}). This contrast is revealing given that all tasks use identical obfuscation statements. Models demonstrate a clear ability to discount irrelevant facts and opinions in simpler tasks, but stumble when the same statements appear in a more complex setting. 

\textit{Reasoning Decomposition}. \textbf{Correct intermediate solutions do not guarantee downstream performance gains and can even degrade it, revealing that reasoning through intermediate steps can be more beneficial than conditioning on partial solutions.} (Figure \ref{figure:rd}, Tables \ref{appendix:rd_efficient} to \ref{appendix:rd_reasoning}). Providing gold intermediate solutions typically boosts downstream performance, indicating that error propagation is a dominant factor in multi-step reasoning failures. Yet exceptions to this trend suggest that partial solutions can disrupt the model’s reasoning trajectory for estate-level tasks that involve long reasoning paths. For example, efficient models often perform worse on Task OE when provided with gold NA solutions, despite NA supplying the exact non-exempt dollar amount that OE seeks to minimize. Notably, four of six reasoning models perform worse on Task NA when provided with gold EC or EV solutions, despite these steps specifying the exemptions and valuation limits needed to compute the total non-exempt amount. This behavior is particularly surprising for two reasons: (1) it is most pronounced in the reasoning models that otherwise exhibit the strongest reasoning capabilities across our experiments (GPT-5, o3, Gemini-2.5-Pro), and (2) it is largely absent in efficient and large non-reasoning models. We hypothesize that reasoning oriented post-training reinforces end-to-end reasoning trajectories, rather than conditional reasoning from partially solved states. As a result, providing intermediate conclusions can sometimes reduce performance by disrupting these reasoning trajectories.

\section{Conclusion}
We introduce OpenExempt, a framework for dynamically generating complex legal tasks grounded in structured legal knowledge, and a diagnostic benchmark for evaluating legal reasoning capabilities in language models. We release OpenExempt to the public to support further research and encourage collaboration between the legal and NLP communities.

\section*{Limitations}

We note several limitations rooted in our design choices. OpenExempt currently: (i) focuses on bankruptcy and state exemption law; (ii) evaluates only U.S. federal law and a small number of selected state jurisdictions; (iii) does not support multilingual tasks; and (iv) focuses on objectively correct tasks, which does not reflect the ambiguity common in legal practice. Given these limitations, OpenExempt should be treated as a complement to current evaluation methods, not a replacement. OpenExempt was designed to be easily extended by either legal or technical skill sets. We believe there is significant potential to build on OpenExempt and view these limitations as natural starting points for future work, including developing and evaluating new approaches to instruction tuning for stepwise legal reasoning.

\section*{Acknowledgments}
This work was supported by the Center for Advancing the Safety of Machine Intelligence (CASMI).


\bibliography{custom}

@misc{wang2024mmluprorobustchallengingmultitask,
      title={MMLU-Pro: A More Robust and Challenging Multi-Task Language Understanding Benchmark}, 
      author={Yubo Wang and Xueguang Ma and Ge Zhang and Yuansheng Ni and Abhranil Chandra and Shiguang Guo and Weiming Ren and Aaran Arulraj and Xuan He and Ziyan Jiang and Tianle Li and Max Ku and Kai Wang and Alex Zhuang and Rongqi Fan and Xiang Yue and Wenhu Chen},
      year={2024},
      eprint={2406.01574},
      archivePrefix={arXiv},
      primaryClass={cs.CL},
      url={https://arxiv.org/abs/2406.01574}, 
}

@misc{alzahrani2024benchmarkstargetsrevealingsensitivity,
      title={When Benchmarks are Targets: Revealing the Sensitivity of Large Language Model Leaderboards}, 
      author={Norah Alzahrani and Hisham Abdullah Alyahya and Yazeed Alnumay and Sultan Alrashed and Shaykhah Alsubaie and Yusef Almushaykeh and Faisal Mirza and Nouf Alotaibi and Nora Altwairesh and Areeb Alowisheq and M Saiful Bari and Haidar Khan},
      year={2024},
      eprint={2402.01781},
      archivePrefix={arXiv},
      primaryClass={cs.CL},
      url={https://arxiv.org/abs/2402.01781}, 
}

@misc{fan2025lexambenchmarkinglegalreasoning,
      title={LEXam: Benchmarking Legal Reasoning on 340 Law Exams}, 
      author={Yu Fan and Jingwei Ni and Jakob Merane and Yang Tian and Yoan Hermstrüwer and Yinya Huang and Mubashara Akhtar and Etienne Salimbeni and Florian Geering and Oliver Dreyer and Daniel Brunner and Markus Leippold and Mrinmaya Sachan and Alexander Stremitzer and Christoph Engel and Elliott Ash and Joel Niklaus},
      year={2025},
      eprint={2505.12864},
      archivePrefix={arXiv},
      primaryClass={cs.CL},
      url={https://arxiv.org/abs/2505.12864}, 
}

@misc{servantez2024chainlogicrulebasedreasoning,
      title={Chain of Logic: Rule-Based Reasoning with Large Language Models}, 
      author={Sergio Servantez and Joe Barrow and Kristian Hammond and Rajiv Jain},
      year={2024},
      eprint={2402.10400},
      archivePrefix={arXiv},
      primaryClass={cs.CL},
      url={https://arxiv.org/abs/2402.10400}, 
}

@misc{blairstanek2023gpt3performstatutoryreasoning,
      title={Can GPT-3 Perform Statutory Reasoning?}, 
      author={Andrew Blair-Stanek and Nils Holzenberger and Benjamin Van Durme},
      year={2023},
      eprint={2302.06100},
      archivePrefix={arXiv},
      primaryClass={cs.CL},
      url={https://arxiv.org/abs/2302.06100}, 
}

@misc{openai2025gpt5,
  author       = {OpenAI},
  title        = {GPT-5 System Card},
  year         = {2025},
  month        = {August},
  day          = {13},
  url          = {https://cdn.openai.com/gpt-5-system-card.pdf},
  note         = {Accessed: 2025-12-25}
}

@misc{openai2025o3o4mini,
  author       = {OpenAI},
  title        = {OpenAI o3 and o4-mini System Card},
  year         = {2025},
  month        = {April},
  day          = {16},
  url          = {https://openai.com/index/o3-o4-mini-system-card/},
  note         = {Accessed: 2025-12-25}
}

@misc{openai2025gpt41,
  author       = {OpenAI},
  title        = {Introducing GPT-4.1 in the API},
  year         = {2025},
  month        = {April},
  day          = {14},
  url          = {https://openai.com/index/gpt-4-1/},
  note         = {Accessed: 2025-12-25}
}

@misc{anthropic2025sonnet4,
  author       = {Anthropic},
  title        = {Claude Sonnet 4: System Card},
  year         = {2025},
  month        = {May},
  url          = {https://www-cdn.anthropic.com/6be99a52cb68eb70eb9572b4cafad13df32ed995.pdf},
  note         = {Accessed: 2025-12-25}
}

@misc{anthropic2024claude3,
  author       = {Anthropic},
  title        = {The Claude 3 Model Family: Opus, Sonnet, Haiku},
  year         = {2024},
  url          = {https://assets.anthropic.com/m/61e7d27f8c8f5919/original/Claude-3-Model-Card.pdf},
  note         = {Accessed: 2025-12-25}
}

@misc{deepseekai2025deepseekr1incentivizingreasoningcapability,
      title={DeepSeek-R1: Incentivizing Reasoning Capability in LLMs via Reinforcement Learning}, 
      author={DeepSeek-AI and Daya Guo and Dejian Yang and Haowei Zhang and Junxiao Song and Ruoyu Zhang and Runxin Xu and Qihao Zhu and Shirong Ma and Peiyi Wang and Xiao Bi and Xiaokang Zhang and Xingkai Yu and Yu Wu and Z. F. Wu and Zhibin Gou and Zhihong Shao and Zhuoshu Li and Ziyi Gao and Aixin Liu and Bing Xue and Bingxuan Wang and Bochao Wu and Bei Feng and Chengda Lu and Chenggang Zhao and Chengqi Deng and Chenyu Zhang and Chong Ruan and Damai Dai and Deli Chen and Dongjie Ji and Erhang Li and Fangyun Lin and Fucong Dai and Fuli Luo and Guangbo Hao and Guanting Chen and Guowei Li and H. Zhang and Han Bao and Hanwei Xu and Haocheng Wang and Honghui Ding and Huajian Xin and Huazuo Gao and Hui Qu and Hui Li and Jianzhong Guo and Jiashi Li and Jiawei Wang and Jingchang Chen and Jingyang Yuan and Junjie Qiu and Junlong Li and J. L. Cai and Jiaqi Ni and Jian Liang and Jin Chen and Kai Dong and Kai Hu and Kaige Gao and Kang Guan and Kexin Huang and Kuai Yu and Lean Wang and Lecong Zhang and Liang Zhao and Litong Wang and Liyue Zhang and Lei Xu and Leyi Xia and Mingchuan Zhang and Minghua Zhang and Minghui Tang and Meng Li and Miaojun Wang and Mingming Li and Ning Tian and Panpan Huang and Peng Zhang and Qiancheng Wang and Qinyu Chen and Qiushi Du and Ruiqi Ge and Ruisong Zhang and Ruizhe Pan and Runji Wang and R. J. Chen and R. L. Jin and Ruyi Chen and Shanghao Lu and Shangyan Zhou and Shanhuang Chen and Shengfeng Ye and Shiyu Wang and Shuiping Yu and Shunfeng Zhou and Shuting Pan and S. S. Li and Shuang Zhou and Shaoqing Wu and Shengfeng Ye and Tao Yun and Tian Pei and Tianyu Sun and T. Wang and Wangding Zeng and Wanjia Zhao and Wen Liu and Wenfeng Liang and Wenjun Gao and Wenqin Yu and Wentao Zhang and W. L. Xiao and Wei An and Xiaodong Liu and Xiaohan Wang and Xiaokang Chen and Xiaotao Nie and Xin Cheng and Xin Liu and Xin Xie and Xingchao Liu and Xinyu Yang and Xinyuan Li and Xuecheng Su and Xuheng Lin and X. Q. Li and Xiangyue Jin and Xiaojin Shen and Xiaosha Chen and Xiaowen Sun and Xiaoxiang Wang and Xinnan Song and Xinyi Zhou and Xianzu Wang and Xinxia Shan and Y. K. Li and Y. Q. Wang and Y. X. Wei and Yang Zhang and Yanhong Xu and Yao Li and Yao Zhao and Yaofeng Sun and Yaohui Wang and Yi Yu and Yichao Zhang and Yifan Shi and Yiliang Xiong and Ying He and Yishi Piao and Yisong Wang and Yixuan Tan and Yiyang Ma and Yiyuan Liu and Yongqiang Guo and Yuan Ou and Yuduan Wang and Yue Gong and Yuheng Zou and Yujia He and Yunfan Xiong and Yuxiang Luo and Yuxiang You and Yuxuan Liu and Yuyang Zhou and Y. X. Zhu and Yanhong Xu and Yanping Huang and Yaohui Li and Yi Zheng and Yuchen Zhu and Yunxian Ma and Ying Tang and Yukun Zha and Yuting Yan and Z. Z. Ren and Zehui Ren and Zhangli Sha and Zhe Fu and Zhean Xu and Zhenda Xie and Zhengyan Zhang and Zhewen Hao and Zhicheng Ma and Zhigang Yan and Zhiyu Wu and Zihui Gu and Zijia Zhu and Zijun Liu and Zilin Li and Ziwei Xie and Ziyang Song and Zizheng Pan and Zhen Huang and Zhipeng Xu and Zhongyu Zhang and Zhen Zhang},
      year={2025},
      eprint={2501.12948},
      archivePrefix={arXiv},
      primaryClass={cs.CL},
      url={https://arxiv.org/abs/2501.12948}, 
}

@misc{google2025geminipro,
  author       = {Google},
  title        = {Gemini 2.5 Pro Model Card},
  year         = {2025},
  month        = {June},
  url          = {https://storage.googleapis.com/deepmind-media/Model-Cards/Gemini-2-5-Pro-Model-Card.pdf},
  note         = {Accessed: 2025-12-25}
}

@misc{google2025geminiflash,
  author       = {Google},
  title        = {Gemini 2.5 Flash Model Card},
  year         = {2025},
  month        = {December},
  url          = {https://storage.googleapis.com/deepmind-media/Model-Cards/Gemini-2-5-Flash-Model-Card.pdf},
  note         = {Accessed: 2025-12-25}
}

@misc{meta2025llama4,
  author       = {Meta},
  title        = {Llama 4 Model Cards and Prompt formats},
  year         = {2025},
  url          = {https://www.llama.com/docs/model-cards-and-prompt-formats/llama4/},
  note         = {Accessed: 2025-12-25}
}

@misc{deepseekai2025deepseekv3technicalreport,
      title={DeepSeek-V3 Technical Report}, 
      author={DeepSeek-AI and Aixin Liu and Bei Feng and Bing Xue and Bingxuan Wang and Bochao Wu and Chengda Lu and Chenggang Zhao and Chengqi Deng and Chenyu Zhang and Chong Ruan and Damai Dai and Daya Guo and Dejian Yang and Deli Chen and Dongjie Ji and Erhang Li and Fangyun Lin and Fucong Dai and Fuli Luo and Guangbo Hao and Guanting Chen and Guowei Li and H. Zhang and Han Bao and Hanwei Xu and Haocheng Wang and Haowei Zhang and Honghui Ding and Huajian Xin and Huazuo Gao and Hui Li and Hui Qu and J. L. Cai and Jian Liang and Jianzhong Guo and Jiaqi Ni and Jiashi Li and Jiawei Wang and Jin Chen and Jingchang Chen and Jingyang Yuan and Junjie Qiu and Junlong Li and Junxiao Song and Kai Dong and Kai Hu and Kaige Gao and Kang Guan and Kexin Huang and Kuai Yu and Lean Wang and Lecong Zhang and Lei Xu and Leyi Xia and Liang Zhao and Litong Wang and Liyue Zhang and Meng Li and Miaojun Wang and Mingchuan Zhang and Minghua Zhang and Minghui Tang and Mingming Li and Ning Tian and Panpan Huang and Peiyi Wang and Peng Zhang and Qiancheng Wang and Qihao Zhu and Qinyu Chen and Qiushi Du and R. J. Chen and R. L. Jin and Ruiqi Ge and Ruisong Zhang and Ruizhe Pan and Runji Wang and Runxin Xu and Ruoyu Zhang and Ruyi Chen and S. S. Li and Shanghao Lu and Shangyan Zhou and Shanhuang Chen and Shaoqing Wu and Shengfeng Ye and Shengfeng Ye and Shirong Ma and Shiyu Wang and Shuang Zhou and Shuiping Yu and Shunfeng Zhou and Shuting Pan and T. Wang and Tao Yun and Tian Pei and Tianyu Sun and W. L. Xiao and Wangding Zeng and Wanjia Zhao and Wei An and Wen Liu and Wenfeng Liang and Wenjun Gao and Wenqin Yu and Wentao Zhang and X. Q. Li and Xiangyue Jin and Xianzu Wang and Xiao Bi and Xiaodong Liu and Xiaohan Wang and Xiaojin Shen and Xiaokang Chen and Xiaokang Zhang and Xiaosha Chen and Xiaotao Nie and Xiaowen Sun and Xiaoxiang Wang and Xin Cheng and Xin Liu and Xin Xie and Xingchao Liu and Xingkai Yu and Xinnan Song and Xinxia Shan and Xinyi Zhou and Xinyu Yang and Xinyuan Li and Xuecheng Su and Xuheng Lin and Y. K. Li and Y. Q. Wang and Y. X. Wei and Y. X. Zhu and Yang Zhang and Yanhong Xu and Yanhong Xu and Yanping Huang and Yao Li and Yao Zhao and Yaofeng Sun and Yaohui Li and Yaohui Wang and Yi Yu and Yi Zheng and Yichao Zhang and Yifan Shi and Yiliang Xiong and Ying He and Ying Tang and Yishi Piao and Yisong Wang and Yixuan Tan and Yiyang Ma and Yiyuan Liu and Yongqiang Guo and Yu Wu and Yuan Ou and Yuchen Zhu and Yuduan Wang and Yue Gong and Yuheng Zou and Yujia He and Yukun Zha and Yunfan Xiong and Yunxian Ma and Yuting Yan and Yuxiang Luo and Yuxiang You and Yuxuan Liu and Yuyang Zhou and Z. F. Wu and Z. Z. Ren and Zehui Ren and Zhangli Sha and Zhe Fu and Zhean Xu and Zhen Huang and Zhen Zhang and Zhenda Xie and Zhengyan Zhang and Zhewen Hao and Zhibin Gou and Zhicheng Ma and Zhigang Yan and Zhihong Shao and Zhipeng Xu and Zhiyu Wu and Zhongyu Zhang and Zhuoshu Li and Zihui Gu and Zijia Zhu and Zijun Liu and Zilin Li and Ziwei Xie and Ziyang Song and Ziyi Gao and Zizheng Pan},
      year={2025},
      eprint={2412.19437},
      archivePrefix={arXiv},
      primaryClass={cs.CL},
      url={https://arxiv.org/abs/2412.19437}, 
}

@misc{gemmateam2025gemma3technicalreport,
      title={Gemma 3 Technical Report}, 
      author={Gemma Team and Aishwarya Kamath and Johan Ferret and Shreya Pathak and Nino Vieillard and Ramona Merhej and Sarah Perrin and Tatiana Matejovicova and Alexandre Ramé and Morgane Rivière and Louis Rouillard and Thomas Mesnard and Geoffrey Cideron and Jean-bastien Grill and Sabela Ramos and Edouard Yvinec and Michelle Casbon and Etienne Pot and Ivo Penchev and Gaël Liu and Francesco Visin and Kathleen Kenealy and Lucas Beyer and Xiaohai Zhai and Anton Tsitsulin and Robert Busa-Fekete and Alex Feng and Noveen Sachdeva and Benjamin Coleman and Yi Gao and Basil Mustafa and Iain Barr and Emilio Parisotto and David Tian and Matan Eyal and Colin Cherry and Jan-Thorsten Peter and Danila Sinopalnikov and Surya Bhupatiraju and Rishabh Agarwal and Mehran Kazemi and Dan Malkin and Ravin Kumar and David Vilar and Idan Brusilovsky and Jiaming Luo and Andreas Steiner and Abe Friesen and Abhanshu Sharma and Abheesht Sharma and Adi Mayrav Gilady and Adrian Goedeckemeyer and Alaa Saade and Alex Feng and Alexander Kolesnikov and Alexei Bendebury and Alvin Abdagic and Amit Vadi and András György and André Susano Pinto and Anil Das and Ankur Bapna and Antoine Miech and Antoine Yang and Antonia Paterson and Ashish Shenoy and Ayan Chakrabarti and Bilal Piot and Bo Wu and Bobak Shahriari and Bryce Petrini and Charlie Chen and Charline Le Lan and Christopher A. Choquette-Choo and CJ Carey and Cormac Brick and Daniel Deutsch and Danielle Eisenbud and Dee Cattle and Derek Cheng and Dimitris Paparas and Divyashree Shivakumar Sreepathihalli and Doug Reid and Dustin Tran and Dustin Zelle and Eric Noland and Erwin Huizenga and Eugene Kharitonov and Frederick Liu and Gagik Amirkhanyan and Glenn Cameron and Hadi Hashemi and Hanna Klimczak-Plucińska and Harman Singh and Harsh Mehta and Harshal Tushar Lehri and Hussein Hazimeh and Ian Ballantyne and Idan Szpektor and Ivan Nardini and Jean Pouget-Abadie and Jetha Chan and Joe Stanton and John Wieting and Jonathan Lai and Jordi Orbay and Joseph Fernandez and Josh Newlan and Ju-yeong Ji and Jyotinder Singh and Kat Black and Kathy Yu and Kevin Hui and Kiran Vodrahalli and Klaus Greff and Linhai Qiu and Marcella Valentine and Marina Coelho and Marvin Ritter and Matt Hoffman and Matthew Watson and Mayank Chaturvedi and Michael Moynihan and Min Ma and Nabila Babar and Natasha Noy and Nathan Byrd and Nick Roy and Nikola Momchev and Nilay Chauhan and Noveen Sachdeva and Oskar Bunyan and Pankil Botarda and Paul Caron and Paul Kishan Rubenstein and Phil Culliton and Philipp Schmid and Pier Giuseppe Sessa and Pingmei Xu and Piotr Stanczyk and Pouya Tafti and Rakesh Shivanna and Renjie Wu and Renke Pan and Reza Rokni and Rob Willoughby and Rohith Vallu and Ryan Mullins and Sammy Jerome and Sara Smoot and Sertan Girgin and Shariq Iqbal and Shashir Reddy and Shruti Sheth and Siim Põder and Sijal Bhatnagar and Sindhu Raghuram Panyam and Sivan Eiger and Susan Zhang and Tianqi Liu and Trevor Yacovone and Tyler Liechty and Uday Kalra and Utku Evci and Vedant Misra and Vincent Roseberry and Vlad Feinberg and Vlad Kolesnikov and Woohyun Han and Woosuk Kwon and Xi Chen and Yinlam Chow and Yuvein Zhu and Zichuan Wei and Zoltan Egyed and Victor Cotruta and Minh Giang and Phoebe Kirk and Anand Rao and Kat Black and Nabila Babar and Jessica Lo and Erica Moreira and Luiz Gustavo Martins and Omar Sanseviero and Lucas Gonzalez and Zach Gleicher and Tris Warkentin and Vahab Mirrokni and Evan Senter and Eli Collins and Joelle Barral and Zoubin Ghahramani and Raia Hadsell and Yossi Matias and D. Sculley and Slav Petrov and Noah Fiedel and Noam Shazeer and Oriol Vinyals and Jeff Dean and Demis Hassabis and Koray Kavukcuoglu and Clement Farabet and Elena Buchatskaya and Jean-Baptiste Alayrac and Rohan Anil and Dmitry and Lepikhin and Sebastian Borgeaud and Olivier Bachem and Armand Joulin and Alek Andreev and Cassidy Hardin and Robert Dadashi and Léonard Hussenot},
      year={2025},
      eprint={2503.19786},
      archivePrefix={arXiv},
      primaryClass={cs.CL},
      url={https://arxiv.org/abs/2503.19786}, 
}

@misc{deepseek2025r1,
  author       = {DeepSeek},
  title        = {DeepSeek R1 Model Card},
  year         = {2025},
  url          = {https://huggingface.co/deepseek-ai/DeepSeek-R1},
  note         = {Accessed: 2025-12-27}
}

@misc{pydantic2025validation,
  author       = {Pydantic},
  title        = {Pydantic Validation},
  year         = {2025},
  url          = {https://docs.pydantic.dev},
  note         = {Accessed: 2025-12-28}
}

@misc{rapidfuzz2025,
  author       = {RapidFuzz},
  title        = {RapidFuzz Documentation},
  year         = {2025},
  url          = {https://rapidfuzz.github.io/RapidFuzz/},
  note         = {Accessed: 2025-12-28}
}

@misc{guha2023legalbenchcollaborativelybuiltbenchmark,
      title={LegalBench: A Collaboratively Built Benchmark for Measuring Legal Reasoning in Large Language Models}, 
      author={Neel Guha and Julian Nyarko and Daniel E. Ho and Christopher Ré and Adam Chilton and Aditya Narayana and Alex Chohlas-Wood and Austin Peters and Brandon Waldon and Daniel N. Rockmore and Diego Zambrano and Dmitry Talisman and Enam Hoque and Faiz Surani and Frank Fagan and Galit Sarfaty and Gregory M. Dickinson and Haggai Porat and Jason Hegland and Jessica Wu and Joe Nudell and Joel Niklaus and John Nay and Jonathan H. Choi and Kevin Tobia and Margaret Hagan and Megan Ma and Michael Livermore and Nikon Rasumov-Rahe and Nils Holzenberger and Noam Kolt and Peter Henderson and Sean Rehaag and Sharad Goel and Shang Gao and Spencer Williams and Sunny Gandhi and Tom Zur and Varun Iyer and Zehua Li},
      year={2023},
      eprint={2308.11462},
      archivePrefix={arXiv},
      primaryClass={cs.CL},
      url={https://arxiv.org/abs/2308.11462}, 
}

@inproceedings{Niklaus_2023,
   title={LEXTREME: A Multi-Lingual and Multi-Task Benchmark for the Legal Domain},
   url={http://dx.doi.org/10.18653/v1/2023.findings-emnlp.200},
   DOI={10.18653/v1/2023.findings-emnlp.200},
   booktitle={Findings of the Association for Computational Linguistics: EMNLP 2023},
   publisher={Association for Computational Linguistics},
   author={Niklaus, Joel and Matoshi, Veton and Rani, Pooja and Galassi, Andrea and Stürmer, Matthias and Chalkidis, Ilias},
   year={2023},
   pages={3016–3054} 
}

@misc{fei2023lawbenchbenchmarkinglegalknowledge,
      title={LawBench: Benchmarking Legal Knowledge of Large Language Models}, 
      author={Zhiwei Fei and Xiaoyu Shen and Dawei Zhu and Fengzhe Zhou and Zhuo Han and Songyang Zhang and Kai Chen and Zongwen Shen and Jidong Ge},
      year={2023},
      eprint={2309.16289},
      archivePrefix={arXiv},
      primaryClass={cs.CL},
      url={https://arxiv.org/abs/2309.16289}, 
}

@misc{chalkidis2022lexgluebenchmarkdatasetlegal,
      title={LexGLUE: A Benchmark Dataset for Legal Language Understanding in English}, 
      author={Ilias Chalkidis and Abhik Jana and Dirk Hartung and Michael Bommarito and Ion Androutsopoulos and Daniel Martin Katz and Nikolaos Aletras},
      year={2022},
      eprint={2110.00976},
      archivePrefix={arXiv},
      primaryClass={cs.CL},
      url={https://arxiv.org/abs/2110.00976}, 
}

@misc{hendrycks2021cuadexpertannotatednlpdataset,
      title={CUAD: An Expert-Annotated NLP Dataset for Legal Contract Review}, 
      author={Dan Hendrycks and Collin Burns and Anya Chen and Spencer Ball},
      year={2021},
      eprint={2103.06268},
      archivePrefix={arXiv},
      primaryClass={cs.CL},
      url={https://arxiv.org/abs/2103.06268}, 
}

@inproceedings{chalkidis-etal-2019-neural,
    title = "Neural Legal Judgment Prediction in {E}nglish",
    author = "Chalkidis, Ilias  and
      Androutsopoulos, Ion  and
      Aletras, Nikolaos",
    editor = "Korhonen, Anna  and
      Traum, David  and
      M{\`a}rquez, Llu{\'i}s",
    booktitle = "Proceedings of the 57th Annual Meeting of the Association for Computational Linguistics",
    month = jul,
    year = "2019",
    address = "Florence, Italy",
    publisher = "Association for Computational Linguistics",
    url = "https://aclanthology.org/P19-1424/",
    doi = "10.18653/v1/P19-1424",
    pages = "4317--4323"
}

@misc{zheng2021doespretraininghelpassessing,
      title={When Does Pretraining Help? Assessing Self-Supervised Learning for Law and the CaseHOLD Dataset}, 
      author={Lucia Zheng and Neel Guha and Brandon R. Anderson and Peter Henderson and Daniel E. Ho},
      year={2021},
      eprint={2104.08671},
      archivePrefix={arXiv},
      primaryClass={cs.CL},
      url={https://arxiv.org/abs/2104.08671}, 
}

@misc{henderson2022pilelawlearningresponsible,
      title={Pile of Law: Learning Responsible Data Filtering from the Law and a 256GB Open-Source Legal Dataset}, 
      author={Peter Henderson and Mark S. Krass and Lucia Zheng and Neel Guha and Christopher D. Manning and Dan Jurafsky and Daniel E. Ho},
      year={2022},
      eprint={2207.00220},
      archivePrefix={arXiv},
      primaryClass={cs.CL},
      url={https://arxiv.org/abs/2207.00220}, 
}

@misc{niklaus2024multilegalpile689gbmultilinguallegal,
      title={MultiLegalPile: A 689GB Multilingual Legal Corpus}, 
      author={Joel Niklaus and Veton Matoshi and Matthias Stürmer and Ilias Chalkidis and Daniel E. Ho},
      year={2024},
      eprint={2306.02069},
      archivePrefix={arXiv},
      primaryClass={cs.CL},
      url={https://arxiv.org/abs/2306.02069}, 
}

@misc{niklaus2025lawinstructresourcestudyinglanguage,
      title={LawInstruct: A Resource for Studying Language Model Adaptation to the Legal Domain}, 
      author={Joel Niklaus and Lucia Zheng and Arya D. McCarthy and Christopher Hahn and Brian M. Rosen and Peter Henderson and Daniel E. Ho and Garrett Honke and Percy Liang and Christopher Manning},
      year={2025},
      eprint={2404.02127},
      archivePrefix={arXiv},
      primaryClass={cs.CL},
      url={https://arxiv.org/abs/2404.02127}, 
}

@misc{zhang2025thinkinglongersmarterevaluating,
      title={Thinking Longer, Not Always Smarter: Evaluating LLM Capabilities in Hierarchical Legal Reasoning}, 
      author={Li Zhang and Matthias Grabmair and Morgan Gray and Kevin Ashley},
      year={2025},
      eprint={2510.08710},
      archivePrefix={arXiv},
      primaryClass={cs.CL},
      url={https://arxiv.org/abs/2510.08710}, 
}

@article{10.1145/3473582,
author = {Merigoux, Denis and Chataing, Nicolas and Protzenko, Jonathan},
title = {Catala: a programming language for the law},
year = {2021},
issue_date = {August 2021},
publisher = {Association for Computing Machinery},
address = {New York, NY, USA},
volume = {5},
number = {ICFP},
url = {https://doi.org/10.1145/3473582},
doi = {10.1145/3473582},
journal = {Proc. ACM Program. Lang.},
month = aug,
articleno = {77},
numpages = {29}
}

@article{Huttner2020CatalaMT,
  title={Catala: Moving towards the future of legal expert systems},
  author={Liane Huttner and Denis Merigoux},
  journal={Artificial Intelligence and Law},
  year={2020},
  url={https://api.semanticscholar.org/CorpusID:225231457}
}

@article{lawskyarticle2022,
author = {Lawsky, Sarah},
year = {2022},
month = {01},
pages = {535},
title = {Coding the Code: Catala and Computationally Accessible Tax Law},
volume = {75},
journal = {SMU Law Review},
doi = {10.25172/smulr.75.3.4}
}

@article{lawskyarticle2017,
author = {Lawsky, Sarah},
year = {2017},
month = {12},
pages = {60-80},
title = {A Logic for Statutes},
volume = {21},
journal = {Florida Tax Review},
doi = {10.5744/ftr.2017.0002}
}

@inproceedings{10.1145/3594536.3595162,
author = {Servantez, Sergio and Lipka, Nedim and Siu, Alexa and Aggarwal, Milan and Krishnamurthy, Balaji and Garimella, Aparna and Hammond, Kristian and Jain, Rajiv},
title = {Computable Contracts by Extracting Obligation Logic Graphs},
year = {2023},
isbn = {9798400701979},
publisher = {Association for Computing Machinery},
address = {New York, NY, USA},
url = {https://doi.org/10.1145/3594536.3595162},
doi = {10.1145/3594536.3595162},
booktitle = {Proceedings of the Nineteenth International Conference on Artificial Intelligence and Law},
pages = {267–276},
numpages = {10},
location = {Braga, Portugal},
series = {ICAIL '23}
}

@misc{holzenberger2020datasetstatutoryreasoningtax,
      title={A Dataset for Statutory Reasoning in Tax Law Entailment and Question Answering}, 
      author={Nils Holzenberger and Andrew Blair-Stanek and Benjamin Van Durme},
      year={2020},
      eprint={2005.05257},
      archivePrefix={arXiv},
      primaryClass={cs.CL},
      url={https://arxiv.org/abs/2005.05257}, 
}

@article{roche2021ergo,
  title={Ergo--a programming language for Smart Legal Contracts},
  author={Roche, Niall and Hernandez, Walter and Chen, Eason and Sim{\'e}on, J{\'e}r{\^o}me and Selman, Dan},
  journal={arXiv preprint arXiv:2112.07064},
  year={2021}
}

@book{cormen2022introduction,
  title     = {Introduction to Algorithms},
  author    = {Cormen, Thomas H. and Leiserson, Charles E. and Rivest, Ronald L. and Stein, Clifford},
  edition   = {4},
  year      = {2022},
  publisher = {MIT Press},
  address   = {Cambridge, MA}
}

@misc{shojaee2025illusionthinkingunderstandingstrengths,
      title={The Illusion of Thinking: Understanding the Strengths and Limitations of Reasoning Models via the Lens of Problem Complexity}, 
      author={Parshin Shojaee and Iman Mirzadeh and Keivan Alizadeh and Maxwell Horton and Samy Bengio and Mehrdad Farajtabar},
      year={2025},
      eprint={2506.06941},
      archivePrefix={arXiv},
      primaryClass={cs.AI},
      url={https://arxiv.org/abs/2506.06941}, 
}

@misc{mirzadeh2025gsmsymbolicunderstandinglimitationsmathematical,
      title={GSM-Symbolic: Understanding the Limitations of Mathematical Reasoning in Large Language Models}, 
      author={Iman Mirzadeh and Keivan Alizadeh and Hooman Shahrokhi and Oncel Tuzel and Samy Bengio and Mehrdad Farajtabar},
      year={2025},
      eprint={2410.05229},
      archivePrefix={arXiv},
      primaryClass={cs.LG},
      url={https://arxiv.org/abs/2410.05229}, 
}

@misc{hofmann2025fluidlanguagemodelbenchmarking,
      title={Fluid Language Model Benchmarking}, 
      author={Valentin Hofmann and David Heineman and Ian Magnusson and Kyle Lo and Jesse Dodge and Maarten Sap and Pang Wei Koh and Chun Wang and Hannaneh Hajishirzi and Noah A. Smith},
      year={2025},
      eprint={2509.11106},
      archivePrefix={arXiv},
      primaryClass={cs.CL},
      url={https://arxiv.org/abs/2509.11106}, 
}

@article{Dahl_2024,
   title={Large Legal Fictions: Profiling Legal Hallucinations in Large Language Models},
   volume={16},
   ISSN={1946-5319},
   url={http://dx.doi.org/10.1093/jla/laae003},
   DOI={10.1093/jla/laae003},
   number={1},
   journal={Journal of Legal Analysis},
   publisher={Oxford University Press (OUP)},
   author={Dahl, Matthew and Magesh, Varun and Suzgun, Mirac and Ho, Daniel E},
   year={2024},
   month=jan, pages={64–93} 
}

@article{surden2012computable,
  title={Computable Contracts},
  author={Surden, Harry},
  journal={UC Davis Law Review},
  volume={46},
  year={2012}
}

@misc{10.1093/oxfordhb/9780198940272.013.0007,
    author = {Guha, Neel and Nyarko, Julian and Ho, Daniel E. and Ré, Christopher},
    isbn = {9780198940272},
    title = {Building GenAI Benchmarks: A Case Study in Legal Applications},
    booktitle = {The Oxford Handbook of the Foundations and Regulation of Generative AI},
    publisher = {Oxford University Press},
    doi = {10.1093/oxfordhb/9780198940272.013.0007},
    url = {https://doi.org/10.1093/oxfordhb/9780198940272.013.0007},
    eprint = {https://academic.oup.com/book/0/chapter/523978823/chapter-ag-pdf/66125429/book_59908_section_523978823.ag.pdf},
}

@misc{clack2017smartcontracttemplatesfoundations,
      title={Smart Contract Templates: foundations, design landscape and research directions}, 
      author={Christopher D. Clack and Vikram A. Bakshi and Lee Braine},
      year={2017},
      eprint={1608.00771},
      archivePrefix={arXiv},
      primaryClass={cs.CY},
      url={https://arxiv.org/abs/1608.00771}, 
}

@inproceedings{zheng2025retrieval, series={CSLAW ’25},
   title={A Reasoning-Focused Legal Retrieval Benchmark},
   url={http://dx.doi.org/10.1145/3709025.3712219},
   DOI={10.1145/3709025.3712219},
   booktitle={Proceedings of the Symposium on Computer Science and Law on ZZZ},
   publisher={ACM},
   author={Zheng, Lucia and Guha, Neel and Arifov, Javokhir and Zhang, Sarah and Skreta, Michal and Manning, Christopher D. and Henderson, Peter and Ho, Daniel E.},
   year={2025},
   month=mar, pages={169–193},
   collection={CSLAW ’25} 
}

@misc{joshi2023ucreatunsupervisedcaseretrieval,
      title={U-CREAT: Unsupervised Case Retrieval using Events extrAcTion}, 
      author={Abhinav Joshi and Akshat Sharma and Sai Kiran Tanikella and Ashutosh Modi},
      year={2023},
      eprint={2307.05260},
      archivePrefix={arXiv},
      primaryClass={cs.IR},
      url={https://arxiv.org/abs/2307.05260}, 
}

\clearpage
\appendix

\section{Appendix} \label{appendix:section}

\subsection{Modular Task Prompts}
\label{appendix:modular_prompts}

We manually write instructions for each task, which are combined with the generated fact patterns and selected exemption statutes to form the task prompts. When the user specifies a task variant (Section \ref{subsection:tasks}), we also include solved intermediate reasoning steps. All task prompt components (instructions, fact patterns, statutes) are stored separately in the benchmark, allowing us to collapse repeated elements across prompts to substantially reduce storage size. This modular design also aligns with our diagnostic evaluation goals by allowing the community to explore changes to question format and phrasing, which prior work has shown can have unpredictable effects on model performance \cite{wang2024mmluprorobustchallengingmultitask, alzahrani2024benchmarkstargetsrevealingsensitivity}. OpenExempt provides a \texttt{TaskDataset} class to handle loading and iterating over examples.

\subsection{Response Format Compliance}
\label{appendix:format_compliance}

\begin{table}[H]
  \centering
  \caption{Frequency and Percentage of Model Responses with Malformed JSON}
  \label{appendix:malformed_response}
  \begin{tabular}{lccccc}
    \toprule
    \textbf{Model} & \textbf{Frequency} & \textbf{Percent} \\
    \midrule
    GPT-5 & 2 & 0.03 \\
    o3 & 4 & 0.05 \\
    o4-mini & 34 & 0.44 \\
    Claude-Sonnet-4 & 1 & 0.01 \\
    Gemini-2.5-Pro & 3 & 0.04 \\
    DeepSeek-R1 & 34 & 0.44 \\
    \midrule
    GPT-4.1 & 12 & 0.15 \\
    Llama-4-Maverick & 173 & 2.22 \\
    DeepSeek-V3 & 3 & 0.04 \\
    \midrule
    Claude-3.5-Haiku & 22 & 0.28 \\
    Gemma-3 & 198 & 2.54 \\
    Gemini-2.5-Flash & 153 & 1.96 \\
    Llama-4-Scout & 117 & 1.50 \\
    \midrule
    Total & 756 & 0.75 \\
    \bottomrule
  \end{tabular}
\end{table}

\subsection{Response Validation} 
\label{appendix:response_validation}

OpenExempt provides an \texttt{Evaluator} class to handle task-specific evaluation logic, including response format compliance and validation of predicted claims. For each sample, the evaluator: (i) isolates the final solution by extracting the suffix after the “FINAL ANSWER:” marker; (ii) parses the response with a task-specific Pydantic parser \cite{pydantic2025validation}; and (iii) normalizes exemption citations (case folding, trimming) and aligns asset descriptions using fuzzy string matching (RapidFuzz \cite{rapidfuzz2025} partial ratio with a threshold of 90) to ensure stable mapping between model predictions and gold labels. For all tasks except OE, evaluation compares predictions directly against the provided gold targets. Since optimal exemption schedules may not be unique, we evaluate Task OE by first validating predicted claims (e.g., ensuring claims obey exemption caps), before comparing against the known optimal outcome. This validation process is grounded in the same symbolic case objects and machine-readable statutes used during task generation. The predicted solution need not match the gold target to be correct, as long as its legally valid and achieves the same degree of protection. 

\onecolumn
\raggedbottom
\clearpage

\subsection{Configuration Parameters}

\begin{table}[H]
  \centering
  \renewcommand{\arraystretch}{1.2}
  \begin{small}
    \setlength{\LTcapwidth}{\linewidth}
    \begin{longtable}{p{0.35\linewidth} p{0.58\linewidth}}
      \caption[OpenExempt Configuration Parameters]{OpenExempt configuration parameters. Each parameter is specified within the configuration file to control task scope and complexity, dataset size, and degree of obfuscation.
      }
      \label{appendix:config}\\
      \toprule
      \textbf{Parameter(s)} & \textbf{Description} \\
      \midrule
      \endfirsthead

      \toprule
      \textbf{Parameter(s)} & \textbf{Description} \\
      \midrule
      \endhead

      \bottomrule
      \endfoot

      \texttt{start\_task\_id}, \texttt{terminal\_task\_id} &
      The process of exempting assets under the Bankruptcy Code proceeds through a fixed sequence of intermediate tasks (see Figure~\ref{figure:tasks}). These configuration parameters specify which portion of that sequence the model is responsible for solving. \texttt{start\_task\_id} marks the first task to be solved, and \texttt{terminal\_task\_id} marks the last. When both are set to the same value (e.g., 3--3), the configuration isolates a single reasoning task; when set to the broadest range (e.g., 1--5), it evaluates the entire exemption process. This design enables fine-grained analysis of how performance changes as cumulative reasoning complexity increases. \\

      \texttt{dataset\_size} &
      Specifies the number of unique tasks, and their corresponding ground-truth solutions, to generate under the given configuration. Each task is independently sampled using the specified asset ranges, jurisdictions, obfuscation settings, and all other configuration parameters. \\

      \texttt{asset\_count\_min}, \texttt{asset\_count\_max} &
      Defines the minimum and maximum number of assets to include in each generated task. The actual asset count is sampled uniformly across this range, ensuring an equal distribution of tasks at each asset count. This allows controlled variation in task complexity across the dataset. \\

      \texttt{married\_ratio} &
      Specifies the proportion of generated tasks that involve married debtors. Marital status affects applicable exemption limits in many jurisdictions. \\

      \texttt{domicile\_count\_min}, \texttt{domicile\_count\_max} &
      The minimum and maximum number of prior domiciles to include in each fact pattern, sampled uniformly across the specified range. Domicile history determines which federal and state exemption laws a Debtor is eligible to claim. \\

      \texttt{state\_jurisdictions} &
      Specifies the set of U.S. state jurisdictions used for task generation. For each task, one jurisdiction is sampled uniformly from this list to serve as the Debtor’s allowable exemption jurisdiction. The exemption statutes for all listed jurisdictions are included in the prompt, requiring the model to identify the correct jurisdiction and apply its exemption laws to the facts. \\

      \texttt{irrelevant\_asset\_facts}, \texttt{irrelevant\_domicile\_facts}, \texttt{asset\_opinions}, \texttt{domicile\_opinions} &
      Boolean parameters that control the inclusion of obfuscating information in the fact pattern. When enabled, the benchmark injects legally immaterial details or subjective statements related to assets or domicile history. These parameters are used to evaluate the model's robustness to distraction, misdirection, and sycophancy by testing its ability to disregard extraneous details while applying the correct legal reasoning. \\

      \texttt{data\_directory}, \texttt{asset\_directory}, \texttt{statute\_directory}, \texttt{template\_directory}, \texttt{output\_directory} &
      File path parameters that specify where the framework loads input resources and saves generated outputs. The input directories point to data dependencies required for task generation (annotated assets, exemption statutes, natural-language templates), while \texttt{output\_directory} designates where generated tasks and solutions are written. \\
      \bottomrule
    \end{longtable}
  \end{small}
\end{table}

\clearpage

\subsection{Benchmark Suite Composition}

\begin{table}[h]
  \centering
  \caption{Summary of configuration settings for each evaluation suite. Each dataset in the benchmark contains a configuration file with the exact construction specification.
  }
  \label{appendix:suite_config}
  \small
  \renewcommand{\arraystretch}{1.3}
  \begin{tabularx}{\textwidth}{l *{6}{>{\centering\arraybackslash}X} c}
    \toprule
    \textbf{Suite} & \textbf{Tasks} & \textbf{Solved Steps} & \textbf{Asset Count} & \textbf{Married Ratio} & \textbf{Domicile Count} & \textbf{Obfusc-ation} & \textbf{States} \\
    \midrule
    Temporal Reasoning & AE & No & N/A & 0.5 & 1-5 & No & All \\
    Reasoning Decomposition & EC-OE & Yes & 6 & 1.0 & 4 & No & WI, IL, OR \\
    Distractor Robustness & All & No & 4 & 0.5 & 3 & Yes & AZ, PA, WI \\
    Sycophancy Robustness & All & No & 4 & 0.5 & 3 & Yes & AZ, PA, WI \\
    Obfuscation Robustness & All & No & 4 & 0.5 & 3 & Yes & AZ, PA, WI \\
    Asset Scaling & EC-OE & No & 2-8 & 0.0 & 2 & No & IL, OR, PA \\
    \midrule
    Basic Competency & All & No & 2 & 0.0 & 2-3 & No & WI, IL \\
    Intermediate Competency & All & No & 3-5 & 0.5 & 4 & Yes & AZ, PA, OR \\
    Advanced Competency & All & No & 6-8 & 1.0 & 5 & Yes & All \\
    \bottomrule
  \end{tabularx}
\end{table}

\subsection{Diagnostic Suite Results}

\begin{table}[h]
  \centering
  \caption{Model Performance (F1) on Temporal Reasoning Suite.}
  \label{appendix:tr}
  \begin{tabular}{lccccc}
    \toprule
    \textbf{Number of Domiciles} & \textbf{One} & \textbf{Two} & \textbf{Three} & \textbf{Four} & \textbf{Five} \\
    \midrule
    GPT-5 & 1.00 & .930 & .853 & .693 & .665 \\
    o3 & 1.00 & .918 & .870 & .703 & .683 \\
    o4-mini & 1.00 & .978 & .888 & .722 & .670 \\
    Claude-Sonnet-4 & 1.00 & .990 & .880 & .735 & .705 \\
    Gemini-2.5-Pro & 1.00 & .990 & .890 & .722 & .682 \\
    DeepSeek-R1 & .993 & .992 & .890 & .715 & .690 \\
    \midrule
    GPT-4.1 & .940 & .968 & .840 & .695 & .657 \\
    Llama-4-Maverick & .943 & .843 & .752 & .595 & .587 \\
    DeepSeek-V3 & .983 & .920 & .821 & .710 & .675 \\
    \midrule
    Claude-3.5-Haiku & .910 & .778 & .579 & .366 & .374 \\
    Gemma-3 & .847 & .637 & .522 & .425 & .415 \\
    Gemini-2.5-Flash & 1.00 & .980 & .850 & .710 & .675 \\
    Llama-4-Scout & .820 & .524 & .468 & .371 & .372 \\
    \bottomrule
  \end{tabular}
\end{table}

\begin{table}[H]
    \caption{Efficient Model Performance (F1) on Reasoning Decomposition Suite.}
    \label{appendix:rd_efficient}
    \begin{tabularx}{\textwidth}{ll *{4}{>{\centering\arraybackslash}X}}
        \toprule
        \thead[l]{\textbf{Task}} & \thead[l]{\textbf{Solved Steps}} & \thead{\textbf{\makecell{Claude-3.5\\Haiku}}} & \thead{\textbf{Gemma-3}} & \thead{\textbf{\makecell{Gemini-2.5\\Flash}}} & \thead{\textbf{\makecell{Llama-4\\Scout}}} \\
        \midrule
        \multirow{2}{*}{EC} 
            & \textit{None} & .402 & .437 & .595 & .267 \\
            & \textit{AE} & .706 & .444 & .854 & .565 \\
        \midrule
        \multirow{3}{*}{EV} 
            & \textit{None} & .326 & .262 & .544 & .223 \\
            & \textit{AE} & .544 & .280 & .712 & .242 \\
            & \textit{EC} & .685 & .492 & .827 & .347 \\
        \midrule
        \multirow{4}{*}{NA} 
            & \textit{None} & .137 & .248 & .513 & .167 \\
            & \textit{AE} & .190 & .187 & .759 & .309 \\
            & \textit{EC} & .322 & .289 & .513 & .297 \\
            & \textit{EV} & .372 & .224 & .462 & .290 \\
        \midrule
        \multirow{5}{*}{OE} 
            & \textit{None} & .113 & .165 & .592 & .148 \\
            & \textit{AE} & .230 & .095 & .718 & .413 \\
            & \textit{EC} & .319 & .319 & .658 & .276 \\
            & \textit{EV} & .361 & .347 & .788 & .291 \\
            & \textit{NA} & .039 & .020 & .507 & .291 \\
        \bottomrule
    \end{tabularx}
\end{table}

\begin{table}[H]
    \caption{Large Model Performance (F1) on Reasoning Decomposition Suite.}
    \label{appendix:rd_large}
    \begin{tabularx}{\textwidth}{ll *{3}{>{\centering\arraybackslash}X}}
        \toprule
        \thead[l]{\textbf{Task}} & \thead[l]{\textbf{Solved Steps}} & \thead{\textbf{GPT-4.1}} & \thead{\textbf{\makecell{Llama-4\\Maverick}}} & \thead{\textbf{\makecell{DeepSeek\\V3}}} \\
        \midrule
        \multirow{2}{*}{EC} 
            & \textit{None} & .265 & .258 & .367 \\
            & \textit{AE} & .629 & .533 & .724 \\
        \midrule
        \multirow{3}{*}{EV} 
            & \textit{None} & .271 & .219 & .396 \\
            & \textit{AE} & .593 & .512 & .559 \\
            & \textit{EC} & .785 & .798 & .750 \\
        \midrule
        \multirow{4}{*}{NA} 
            & \textit{None} & .481 & .197 & .288 \\
            & \textit{AE} & .511 & .445 & .473 \\
            & \textit{EC} & .361 & .420 & .297 \\
            & \textit{EV} & .441 & .503 & .397 \\
        \midrule
        \multirow{5}{*}{OE} 
            & \textit{None} & .305 & .214 & .387 \\
            & \textit{AE} & .519 & .496 & .551 \\
            & \textit{EC} & .561 & .473 & .462 \\
            & \textit{EV} & .773 & .473 & .571 \\
            & \textit{NA} & .450 & .305 & .473 \\
        \bottomrule
    \end{tabularx}
\end{table}

\begin{table}[H]
    \caption{Reasoning Model Performance (F1) on Reasoning Decomposition Suite.}
    \label{appendix:rd_reasoning}
    \begin{tabularx}{\textwidth}{ll *{6}{>{\centering\arraybackslash}X}}
        \toprule
        \thead[l]{\textbf{Task}} & \thead[l]{\textbf{Solved Steps}} & \thead{\textbf{GPT-5}} & \thead{\textbf{o3}} & \thead{\textbf{o4-mini}} & \thead{\textbf{Sonnet-4}} & \thead{\textbf{\makecell{Gemini \\ Pro}}} & \thead{\textbf{\makecell{DeepSeek \\ R1}}} \\
        \midrule
        \multirow{2}{*}{EC} 
            & \textit{None} & .714 & .742 & .499 & .534 & .714 & .575 \\
            & \textit{AE} & .983 & .983 & .789 & .924 & .958 & .803 \\
        \midrule
        \multirow{3}{*}{EV} 
            & \textit{None} & .671 & .700 & .549 & .402 & .668 & .552 \\
            & \textit{AE} & .800 & .814 & .695 & .716 & .811 & .610 \\
            & \textit{EC} & .836 & .875 & .882 & .869 & .851 & .847 \\
        \midrule
        \multirow{4}{*}{NA} 
            & \textit{None} & .650 & .681 & .427 & .414 & .662 & .582 \\
            & \textit{AE} & .917 & .907 & .649 & .735 & .840 & .709 \\
            & \textit{EC} & .588 & .529 & .524 & .519 & .567 & .499 \\
            & \textit{EV} & .557 & .539 & .407 & .531 & .563 & .458 \\
        \midrule
        \multirow{5}{*}{OE} 
            & \textit{None} & .611 & .630 & .485 & .462 & .684 & .540 \\
            & \textit{AE} & .788 & .765 & .621 & .658 & .788 & .649 \\
            & \textit{EC} & .780 & .734 & .658 & .658 & .857 & .601 \\
            & \textit{EV} & .876 & .795 & .649 & .726 & .876 & .693 \\
            & \textit{NA} & .658 & .693 & .701 & .621 & .649 & .649 \\
        \bottomrule
    \end{tabularx}
\end{table}

\begin{table}[H]
    \caption{Efficient Model Performance (F1) on Baseline, Distractor, Sycophancy, and Obfuscation Robustness Suites.}
    \label{appendix:or_efficient}
    \begin{tabularx}{\textwidth}{ll *{4}{>{\centering\arraybackslash}X}}
        \toprule
        \thead[l]{\textbf{Task}} & \thead[l]{\textbf{Suite}} & \thead{\textbf{\makecell{Claude-3.5\\Haiku}}} & \thead{\textbf{Gemma-3}} & \thead{\textbf{\makecell{Gemini-2.5\\Flash}}} & \thead{\textbf{\makecell{Llama-4\\Scout}}} \\
        \midrule
        \multirow{4}{*}{AE} 
            & \textit{Baseline} & .618 & .528 & .745 & .577 \\
            & \textit{Distractor} & .645 & .575 & .831 & .510 \\
            & \textit{Sycophancy} & .507 & .377 & .831 & .442 \\
            & \textit{Obfuscation} & .509 & .390 & .796 & .457 \\
        \midrule
        \multirow{4}{*}{EC} 
            & \textit{Baseline} & .478 & .466 & .772 & .394 \\
            & \textit{Distractor} & .453 & .477 & .739 & .459 \\
            & \textit{Sycophancy} & .312 & .385 & .749 & .314 \\
            & \textit{Obfuscation} & .319 & .359 & .719 & .390 \\
        \midrule
        \multirow{4}{*}{EV} 
            & \textit{Baseline} & .382 & .264 & .715 & .333 \\
            & \textit{Distractor} & .378 & .308 & .677 & .298 \\
            & \textit{Sycophancy} & .236 & .199 & .701 & .271 \\
            & \textit{Obfuscation} & .262 & .201 & .652 & .237 \\
        \midrule
        \multirow{4}{*}{NA} 
            & \textit{Baseline} & .297 & .356 & .763 & .316 \\
            & \textit{Distractor} & .154 & .189 & .753 & .194 \\
            & \textit{Sycophancy} & .159 & .198 & .776 & .152 \\
            & \textit{Obfuscation} & .112 & .115 & .787 & .130 \\
        \midrule
        \multirow{4}{*}{OE} 
            & \textit{Baseline} & .276 & .400 & .817 & .387 \\
            & \textit{Distractor} & .058 & .000 & .621 & .020 \\
            & \textit{Sycophancy} & .165 & .165 & .718 & .131 \\
            & \textit{Obfuscation} & .039 & .000 & .582 & .000 \\
        \bottomrule
    \end{tabularx}
\end{table}

\begin{table}[H]
    \caption{Large Model Performance (F1) on Baseline, Distractor, Sycophancy, and Obfuscation Robustness Suites.}
    \label{appendix:or_large}
    \begin{tabularx}{\textwidth}{ll *{3}{>{\centering\arraybackslash}X}}
        \toprule
        \thead[l]{\textbf{Task}} & \thead[l]{\textbf{Suite}} & \thead{\textbf{GPT-4.1}} & \thead{\textbf{\makecell{Llama-4\\Maverick}}} & \thead{\textbf{\makecell{DeepSeek\\V3}}} \\
        \midrule
        \multirow{4}{*}{AE} 
            & \textit{Baseline} & .837 & .820 & .813 \\
            & \textit{Distractor} & .835 & .720 & .802 \\
            & \textit{Sycophancy} & .837 & .748 & .764 \\
            & \textit{Obfuscation} & .788 & .755 & .710 \\
        \midrule
        \multirow{4}{*}{EC} 
            & \textit{Baseline} & .313 & .341 & .478 \\
            & \textit{Distractor} & .335 & .408 & .472 \\
            & \textit{Sycophancy} & .143 & .303 & .219 \\
            & \textit{Obfuscation} & .095 & .277 & .239 \\
        \midrule
        \multirow{4}{*}{EV} 
            & \textit{Baseline} & .285 & .369 & .407 \\
            & \textit{Distractor} & .273 & .355 & .401 \\
            & \textit{Sycophancy} & .118 & .232 & .208 \\
            & \textit{Obfuscation} & .105 & .192 & .195 \\
        \midrule
        \multirow{4}{*}{NA} 
            & \textit{Baseline} & .555 & .416 & .461 \\
            & \textit{Distractor} & .467 & .271 & .328 \\
            & \textit{Sycophancy} & .517 & .321 & .198 \\
            & \textit{Obfuscation} & .450 & .189 & .186 \\
        \midrule
        \multirow{4}{*}{OE} 
            & \textit{Baseline} & .601 & .413 & .507 \\
            & \textit{Distractor} & .462 & .131 & .214 \\
            & \textit{Sycophancy} & .246 & .374 & .246 \\
            & \textit{Obfuscation} & .165 & .113 & .113 \\
        \bottomrule
    \end{tabularx}
\end{table}

\begin{table}[H]
    \caption{Reasoning Model Performance (F1) on Baseline, Distractor, Sycophancy, and Obfuscation Robustness Suites.}
    \label{appendix:or_reasoning}
    \begin{tabularx}{\textwidth}{ll *{6}{>{\centering\arraybackslash}X}}
        \toprule
        \thead[l]{\textbf{Task}} & \thead[l]{\textbf{Suite}} & \thead{\textbf{GPT-5}} & \thead{\textbf{o3}} & \thead{\textbf{o4-mini}} & \thead{\textbf{Sonnet-4}} & \thead{\textbf{\makecell{Gemini \\ Pro}}} & \thead{\textbf{\makecell{DeepSeek \\ R1}}} \\
        \midrule
        \multirow{4}{*}{AE} 
            & \textit{Baseline} & .785 & .808 & .825 & .810 & .835 & .835 \\
            & \textit{Distractor} & .843 & .845 & .855 & .865 & .895 & .887 \\
            & \textit{Sycophancy} & .792 & .805 & .823 & .830 & .843 & .827 \\
            & \textit{Obfuscation} & .820 & .827 & .842 & .820 & .852 & .830 \\
        \midrule
        \multirow{4}{*}{EC} 
            & \textit{Baseline} & .824 & .816 & .725 & .597 & .840 & .755 \\
            & \textit{Distractor} & .837 & .832 & .636 & .535 & .839 & .738 \\
            & \textit{Sycophancy} & .858 & .871 & .528 & .486 & .833 & .533 \\
            & \textit{Obfuscation} & .779 & .828 & .469 & .419 & .795 & .435 \\
        \midrule
        \multirow{4}{*}{EV} 
            & \textit{Baseline} & .771 & .737 & .602 & .532 & .767 & .637 \\
            & \textit{Distractor} & .755 & .734 & .590 & .487 & .784 & .656 \\
            & \textit{Sycophancy} & .804 & .787 & .456 & .411 & .760 & .510 \\
            & \textit{Obfuscation} & .792 & .778 & .468 & .336 & .759 & .432 \\
        \midrule
        \multirow{4}{*}{NA} 
            & \textit{Baseline} & .862 & .868 & .644 & .511 & .865 & .789 \\
            & \textit{Distractor} & .850 & .764 & .573 & .557 & .837 & .705 \\
            & \textit{Sycophancy} & .867 & .835 & .560 & .539 & .832 & .818 \\
            & \textit{Obfuscation} & .826 & .727 & .467 & .474 & .827 & .674 \\
        \midrule
        \multirow{4}{*}{OE} 
            & \textit{Baseline} & .844 & .876 & .734 & .611 & .870 & .817 \\
            & \textit{Distractor} & .529 & .148 & .400 & .305 & .639 & .507 \\
            & \textit{Sycophancy} & .742 & .765 & .621 & .450 & .758 & .742 \\
            & \textit{Obfuscation} & .450 & .182 & .305 & .347 & .701 & .291 \\
        \bottomrule
    \end{tabularx}
\end{table}

\begin{table}[H]
    \caption{Efficient Model Performance (F1) on Asset Scaling Suite.}
    \label{appendix:as_efficient}
    \begin{tabularx}{\textwidth}{ll *{4}{>{\centering\arraybackslash}X}}
        \toprule
        \thead[l]{\textbf{Task}} & \thead[l]{\textbf{\makecell[l]{Asset\\Count}}} & \thead{\textbf{\makecell{Claude-3.5\\Haiku}}} & \thead{\textbf{Gemma-3}} & \thead{\textbf{\makecell{Gemini-2.5\\Flash}}} & \thead{\textbf{\makecell{Llama-4\\Scout}}} \\
        \midrule
        \multirow{4}{*}{EC} 
            & \textit{2} & .484 & .445 & .927 & .426 \\
            & \textit{4} & .452 & .420 & .902 & .382 \\
            & \textit{6} & .432 & .404 & .898 & .367 \\
            & \textit{8} & .412 & .434 & .888 & .369 \\
        \midrule
        \multirow{4}{*}{EV} 
            & \textit{2} & .338 & .362 & .910 & .367 \\
            & \textit{4} & .343 & .319 & .899 & .334 \\
            & \textit{6} & .318 & .235 & .892 & .270 \\
            & \textit{8} & .328 & .254 & .866 & .314 \\
        \midrule
        \multirow{4}{*}{NA} 
            & \textit{2} & .411 & .370 & .868 & .416 \\
            & \textit{4} & .314 & .276 & .848 & .354 \\
            & \textit{6} & .191 & .242 & .832 & .290 \\
            & \textit{8} & .140 & .266 & .845 & .239 \\
        \midrule
        \multirow{4}{*}{OE} 
            & \textit{2} & .462 & .551 & .942 & .425 \\
            & \textit{4} & .131 & .148 & .883 & .230 \\
            & \textit{6} & .058 & .182 & .817 & .131 \\
            & \textit{8} & .020 & .077 & .810 & .077 \\
        \bottomrule
    \end{tabularx}
\end{table}

\begin{table}[H]
    \caption{Large Model Performance (F1) on Asset Scaling Suite.}
    \label{appendix:as_large}
    \begin{tabularx}{\textwidth}{ll *{3}{>{\centering\arraybackslash}X}}
        \toprule
        \thead[l]{\textbf{Task}} & \thead[l]{\textbf{\makecell[l]{Asset\\Count}}} & \thead{\textbf{GPT-4.1}} & \thead{\textbf{\makecell{Llama-4\\Maverick}}} & \thead{\textbf{\makecell{DeepSeek\\V3}}} \\
        \midrule
        \multirow{4}{*}{EC} 
            & \textit{2} & .441 & .514 & .473 \\
            & \textit{4} & .345 & .465 & .521 \\
            & \textit{6} & .363 & .479 & .520 \\
            & \textit{8} & .327 & .470 & .439 \\
        \midrule
        \multirow{4}{*}{EV} 
            & \textit{2} & .437 & .447 & .426 \\
            & \textit{4} & .416 & .438 & .477 \\
            & \textit{6} & .349 & .437 & .457 \\
            & \textit{8} & .364 & .447 & .429 \\
        \midrule
        \multirow{4}{*}{NA} 
            & \textit{2} & .588 & .528 & .562 \\
            & \textit{4} & .545 & .480 & .444 \\
            & \textit{6} & .546 & .472 & .483 \\
            & \textit{8} & .531 & .433 & .411 \\
        \midrule
        \multirow{4}{*}{OE} 
            & \textit{2} & .611 & .630 & .667 \\
            & \textit{4} & .462 & .450 & .387 \\
            & \textit{6} & .276 & .333 & .361 \\
            & \textit{8} & .182 & .182 & .182 \\
        \bottomrule
    \end{tabularx}
\end{table}

\begin{table}[H]
    \caption{Reasoning Model Performance (F1) on Asset Scaling Suite.}
    \label{appendix:as_reasoning}
    \begin{tabularx}{\textwidth}{ll *{6}{>{\centering\arraybackslash}X}}
        \toprule
        \thead[l]{\textbf{Task}} & \thead[l]{\textbf{\makecell[l]{Asset\\Count}}} & \thead{\textbf{GPT-5}} & \thead{\textbf{o3}} & \thead{\textbf{o4-mini}} & \thead{\textbf{Sonnet-4}} & \thead{\textbf{\makecell{Gemini \\ Pro}}} & \thead{\textbf{\makecell{DeepSeek \\ R1}}} \\
        \midrule
        \multirow{4}{*}{EC} 
            & \textit{2} & .964 & .968 & .794 & .807 & .951 & .809 \\
            & \textit{4} & .981 & .961 & .720 & .770 & .950 & .801 \\
            & \textit{6} & .941 & .965 & .754 & .712 & .958 & .802 \\
            & \textit{8} & .941 & .945 & .701 & .706 & .935 & .804 \\
        \midrule
        \multirow{4}{*}{EV} 
            & \textit{2} & .949 & .956 & .763 & .736 & .934 & .774 \\
            & \textit{4} & .968 & .945 & .737 & .721 & .946 & .752 \\
            & \textit{6} & .957 & .918 & .720 & .658 & .946 & .742 \\
            & \textit{8} & .955 & .950 & .683 & .666 & .935 & .729 \\
        \midrule
        \multirow{4}{*}{NA} 
            & \textit{2} & .927 & .955 & .737 & .676 & .883 & .870 \\
            & \textit{4} & .898 & .962 & .734 & .633 & .898 & .850 \\
            & \textit{6} & .918 & .942 & .687 & .605 & .889 & .870 \\
            & \textit{8} & .938 & .949 & .717 & .643 & .897 & .867 \\
        \midrule
        \multirow{4}{*}{OE} 
            & \textit{2} & .947 & .995 & .824 & .795 & .953 & .919 \\
            & \textit{4} & .936 & .953 & .693 & .675 & .969 & .837 \\
            & \textit{6} & .901 & .913 & .658 & .571 & .925 & .734 \\
            & \textit{8} & .913 & .901 & .485 & .496 & .919 & .667 \\
        \bottomrule
    \end{tabularx}
\end{table}

\subsection{Task Prompt Examples}
\label{appendix:task_examples}

{
\small
\begin{tcolorbox}[enhanced,
  attach boxed title to top center={yshift=-3mm,yshifttext=-1mm},
  colback=white,
  colframe=blue!50!black,
  colbacktitle=blue!50!black,
  title=Temporal Reasoning Suite: Task AE,
  fonttitle=\bfseries
  ]
    Determine which state or federal exemptions may be claimed by the Debtor(s) under the provided statutes.
    \\\\
    Your answer to this task must be based solely on applying the provided Federal and State statutes to the given facts.
    \\\\
    Response Format: Your response must end with your final answer in the following template: FINAL ANSWER: [YOUR FINAL ANSWER]. Your final answer must consist of only a comma-separated list of jurisdictions, without any additional text. States should be identified by name. If federal exemptions are allowed, include 'Federal' in the list. Example response format: 
    FINAL ANSWER: Alaska, Federal
    \\\\
    Facts:
    Luis Gonzalez (hereinafter the Debtor) filed for bankruptcy on 14 July 2024. After moving their household to Delta, Pennsylvania on Saturday, March 21st, 2020, Luis Gonzalez eventually relocated to Marana, Arizona on 29th of February 2024.
    \\\\
    Statutes:
\end{tcolorbox}
}

{
\small
\begin{tcolorbox}[enhanced,
  attach boxed title to top center={yshift=-3mm,yshifttext=-1mm},
  colback=white,
  colframe=green!50!black,
  colbacktitle=green!50!black,
  title=Solution,
  fonttitle=\bfseries
  ]
  Federal, Pennsylvania
\end{tcolorbox}
}

{
\small
\begin{tcolorbox}[enhanced,
  attach boxed title to top center={yshift=-3mm,yshifttext=-1mm},
  colback=white,
  colframe=blue!50!black,
  colbacktitle=blue!50!black,
  title=Distractor Robustness Suite: Task EC,
  fonttitle=\bfseries
  ]
    For each asset in the estate, identify all applicable exemptions under which that asset may be protected.
    \\\\
    Your answer to this task must be based solely on applying the provided Federal and State statutes to the given facts. If the task involves a married couple, assume all assets mentioned are jointly owned, with each spouse holding an equal undivided interest, unless explicitly stated otherwise. Assume all assets are held for the personal use of the Debtor(s), unless explicitly stated otherwise.
    \\\\
    Response Format: Your response must end with your final answer in the following template: FINAL ANSWER: [YOUR FINAL ANSWER]. Your final answer must consist of only valid JSON in the exact format specified below. Provide your final answer as a JSON object where: each key is the exact asset description provided in the fact pattern, and each value is an array of applicable exemption citations. Example response format: 
    FINAL ANSWER: {"1981 DeLorean DMC-12": ["11 U.S.C. § 522(d)(2)", "11 U.S.C. § 522(d)(5)"]}
    \\\\
    Facts:
    Megha and Dalia Joshi (hereinafter the Debtors) filed for bankruptcy on March 7th 2024. The Debtors began living in Litchfield Park, Arizona on 19 March 2012, but relocated to Waunakee, Wisconsin on 9.10.2021. For the 28 days following that date, Megha stayed in Patagonia, Arizona to complete a boater safety course and obtain a state-issued boating certificate, a new requirement for their job as a marine biologist. While there, they signed a rental agreement (Contract \#R-781) for a boat slip at a local marina to berth the assigned training vessel for the duration of the course. The Joshis chose to relocate their household to Weyauwega, Wisconsin on 12th of September, 2021. Records show that Megha and Dalia Joshi possess a compact Bluetooth speaker with splash-resistant casing with a value of \$400.00. Megha and Dalia own a calico cat worth \$145.00. The Debtor's name is listed on a UTMA savings account, and the corresponding Form 1099-INT is mailed to their address. The account holds a balance of \$8,150, which originated as an irrevocable gift from the Debtor's brother to the account's beneficiary, the Debtor's 14 year-old nephew. The known assets of Megha and Dalia Joshi include a woven tapestry wall hanging with bohemian motif worth \$1,305.00 and a Hi-Point C9 9mm pistol appraised at \$195.00.
    \\\\
    Statutes:
\end{tcolorbox}
}

{
\small
\begin{tcolorbox}[enhanced,
  attach boxed title to top center={yshift=-3mm,yshifttext=-1mm},
  colback=white,
  colframe=green!50!black,
  colbacktitle=green!50!black,
  title=Solution,
  fonttitle=\bfseries
  ]
    \{"compact Bluetooth speaker with splash-resistant casing": ["11 U.S.C. § 522(d)(3)", "11 U.S.C. § 522(d)(5)", "Wis. Stat. § 815.18(3)(d)"], "calico cat": ["11 U.S.C. § 522(d)(3)", "11 U.S.C. § 522(d)(5)", "Wis. Stat. § 815.18(3)(d)"], "woven tapestry wall hanging with bohemian motif": ["11 U.S.C. § 522(d)(3)", "11 U.S.C. § 522(d)(5)", "Wis. Stat. § 815.18(3)(d)"], "Hi-Point C9 9mm pistol": ["11 U.S.C. § 522(d)(5)", "Wis. Stat. § 815.18(3)(d)"]\}
\end{tcolorbox}
}

{
\small
\begin{tcolorbox}[enhanced,
  attach boxed title to top center={yshift=-3mm,yshifttext=-1mm},
  colback=white,
  colframe=blue!50!black,
  colbacktitle=blue!50!black,
  title=Reasoning Decomposition Suite: Task OE,
  fonttitle=\bfseries
  ]
    Determine the optimal set of exemptions to best protect the assets in the estate. The goal is to minimize the total dollar value of non-exempt assets. If property may be exempted under multiple jurisdictions, you must select the jurisdiction that would result in the best solution.
    \\\\
    Your answer to this task must be based solely on applying the provided Federal and State statutes to the given facts. If the task involves a married couple, assume all assets mentioned are jointly owned, with each spouse holding an equal undivided interest, unless explicitly stated otherwise. Assume all assets are held for the personal use of the Debtor(s), unless explicitly stated otherwise.
    \\\\
    Response Format: Your response must end with your final answer in the following template: FINAL ANSWER: [YOUR FINAL ANSWER]. Your final answer must consist of only valid JSON in the exact format specified below. Provide your response as a JSON object where: each key is the exact asset description provided in the fact pattern, and each value is an array representing the optimal exemptions for this asset. Each exemption in this array is an object containing a citation and a claim value. Claim values must not contain any commas or dollar signs. Example response format: 
    FINAL ANSWER: {"1981 DeLorean DMC-12": [{"citation": "11 U.S.C. § 522(d)(2)", "claim\_value": 3000}]}
    \\\\
    Facts:
    Tobias and Leon Fischer (hereinafter the Debtors) filed for bankruptcy on 10th day of January, 2024. Ownership of the small mountain cabin used year-round as the principal residence, priced at \$49,500.00, is claimed by Tobias and Leon Fischer. The pair of suede ankle boots with zipper closure currently owned by the Fischers carries a value of \$225.00. The Debtors assert ownership of an audiologist prescribed custom-fit hearing aids with behind-the-ear receiver and noise filtering with a market value of \$1,425.00. A 14-karat gold engagement band with engraving on the inner surface valued at \$770.00 is under the ownership of Tobias and Leon. An oxygen concentrator with portable carry cart and backup battery (physician authorized) appraised at \$3,250.00 is the property of Tobias and Leon Fischer. A disclosure of assets by the Fischers reports a floor-length curtains with floral embroidery with a current value of \$280.00.
    \\\\
    Solved Reasoning Steps:
    The following reasoning steps have already been solved. Use this information to aid you in completing the remainder of the task.
    All applicable exemptions have been identified below for each asset in the estate.
    The small mountain cabin used year-round as the principal residence may be exempted under 11 U.S.C. § 522(d)(1), 11 U.S.C. § 522(d)(5), and Or. Rev. Stat. § 18.395(1).
    The pair of suede ankle boots with zipper closure may be exempted under 11 U.S.C. § 522(d)(3), 11 U.S.C. § 522(d)(5), Or. Rev. Stat. § 18.345(1)(b), and Or. Rev. Stat. § 18.345(1)(p).
    The audiologist prescribed custom-fit hearing aids with behind-the-ear receiver and noise filtering may be exempted under 11 U.S.C. § 522(d)(5), 11 U.S.C. § 522(d)(9), Or. Rev. Stat. § 18.345(1)(h), and Or. Rev. Stat. § 18.345(1)(p).
    The 14-karat gold engagement band with engraving on the inner surface may be exempted under 11 U.S.C. § 522(d)(4), 11 U.S.C. § 522(d)(5), Or. Rev. Stat. § 18.345(1)(b), and Or. Rev. Stat. § 18.345(1)(p).
    The oxygen concentrator with portable carry cart and backup battery (physician authorized) may be exempted under 11 U.S.C. § 522(d)(5), 11 U.S.C. § 522(d)(9), Or. Rev. Stat. § 18.345(1)(h), and Or. Rev. Stat. § 18.345(1)(p).
    The floor-length curtains with floral embroidery may be exempted under 11 U.S.C. § 522(d)(3), 11 U.S.C. § 522(d)(5), Or. Rev. Stat. § 18.345(1)(f), and Or. Rev. Stat. § 18.345(1)(p).
    \\\\
    Statutes:
\end{tcolorbox}
}

{
\small
\begin{tcolorbox}[enhanced,
  attach boxed title to top center={yshift=-3mm,yshifttext=-1mm},
  colback=white,
  colframe=green!50!black,
  colbacktitle=green!50!black,
  title=Solution,
  fonttitle=\bfseries
  ]
    \{"small mountain cabin used year-round as the principal residence": [\{"citation": "11 U.S.C. § 522(d)(5)", "claim\_value": 30850\}, \{"citation": "11 U.S.C. § 522(d)(1)", "claim\_value": 18650\}], "pair of suede ankle boots with zipper closure": [\{"citation": "11 U.S.C. § 522(d)(3)", "claim\_value": 225\}], "audiologist prescribed custom-fit hearing aids with behind-the-ear receiver and noise filtering": [\{"citation": "11 U.S.C. § 522(d)(9)", "claim\_value": 1425\}], "14-karat gold engagement band with engraving on the inner surface": [\{"citation": "11 U.S.C. § 522(d)(4)", "claim\_value": 770\}], "oxygen concentrator with portable carry cart and backup battery (physician authorized)": [\{"citation": "11 U.S.C. § 522(d)(9)", "claim\_value": 3250\}], "floor-length curtains with floral embroidery": [\{"citation": "11 U.S.C. § 522(d)(3)", "claim\_value": 280\}]\}
\end{tcolorbox}
}

\subsection{Statute Source by Jurisdiction}
\label{appendix:statute_sources}

\begin{tabularx}{\textwidth}{l X}
    \toprule
    \textbf{Jurisdiction} & \textbf{Source} \\
    \midrule
    Federal & \url{https://uscode.house.gov} \\
    \addlinespace
    Arizona & \url{https://www.azleg.gov/arstitle/} \\
    \addlinespace
    Illinois & \url{https://www.ilga.gov/legislation/ilcs/ilcs.asp} \\
    \addlinespace
    Oregon & \url{https://www.oregonlegislature.gov/bills_laws/pages/ors.aspx} \\
    \addlinespace
    Pennsylvania & \url{https://www.palegis.us/statutes/consolidated} \\
    \addlinespace
    Wisconsin & \url{https://docs.legis.wisconsin.gov/statutes/statutes/815} \\
    \bottomrule
\end{tabularx}

\end{document}